\documentclass[11pt]{article}

\usepackage[preprint]{acl}

\usepackage{times}
\usepackage{latexsym}

\usepackage[T1]{fontenc}

\usepackage[utf8]{inputenc}
 
\usepackage{microtype}

\usepackage{inconsolata}

\usepackage{enumitem}     
\usepackage{mdwlist}      

\usepackage{booktabs}
\usepackage{multirow}
\usepackage{graphicx} 
\usepackage{makecell} 
\usepackage{xcolor}
\usepackage{amsmath}
\usepackage{amsfonts}
\usepackage{pifont}
\newcommand{\cmark}{\ding{51}}
\newcommand{\xmark}{\ding{55}}
\usepackage[most]{tcolorbox}
\usepackage{xcolor}
\usepackage{fvextra}
\usepackage{tabularx}
\usepackage{needspace}

\title{\textsc{UniQL}: Towards Dialect-Universal Benchmarking for Text-to-SQL}

\author{
\textbf{Jianling Gao\textsuperscript{1}},
\textbf{Chongyang Tao\textsuperscript{1}},
\textbf{Jiayuan Bai\textsuperscript{1}},
\textbf{Liu Yang\textsuperscript{1}} 
\textbf{Xuanguang Pan\textsuperscript{1}},
\\
\textbf{Jinrui Liu\textsuperscript{1}},
\textbf{Shihao Xing\textsuperscript},
\textbf{Xiaohan Xu\textsuperscript{2}},
\textbf{Jie Liang\textsuperscript{1}},
\textbf{Shuai Ma\textsuperscript{1}}
\\[1ex]
\textsuperscript{1}SKLCCSE, Beihang University \\
\textsuperscript{2}The University of Hong Kong \\[1ex]
\small{
\textsuperscript{1}\texttt{\{jianlingg,chongyang,baijiayuan,panxg,mashuai\}@buaa.edu.cn}
} \\
\small{
\textsuperscript{2}\texttt{shawnxxh@gmail.com}
}
}
\begin{document}
\maketitle
\begin{abstract}
Existing text-to-SQL benchmarks are largely centered on SQLite, making it difficult to evaluate whether models can generalize across heterogeneous SQL dialects. However, real-world database systems differ substantially in syntax, functions, type systems, and execution semantics, so the same natural language intent often requires dialect-specific SQL realizations. We introduce \textsc{UniQL}, a human-verified benchmark for cross-dialect text-to-SQL evaluation. \textsc{UniQL} aligns 1,534 natural language questions with executable SQL annotations across 16 SQL dialects, yielding 24,544 dialect-specific queries. All dialects share the same intents, aligned schemas and database contents, enabling controlled evaluation of dialect generalization. \textsc{UniQL} is constructed through a hybrid pipeline combining database migration, SQL translation, execution-guided verification, iterative rule summarization, and human validation. Experiments on both open-source and closed-source LLMs show that current models remain far from dialect-universal, with substantial performance variation across database systems and limited transfer from SQLite success to other dialects. These findings highlight the need for aligned cross-dialect benchmarks and more dialect-aware text-to-SQL methods. Code and data are available at \href{https://github.com/JerryGao818/UniQL}{this link}.

\end{abstract}

\section{Introduction}






Structured query languages remain the primary interface for interacting with modern database systems. Despite decades of standardization efforts, the SQL ecosystem is still highly fragmented across database engines such as PostgreSQL, MySQL, Oracle, etc. Although these systems all adopt SQL-like interfaces, they often differ substantially in syntax, built-in functions, and optimization behaviors. As a result, semantically equivalent queries frequently require dialect-specific implementations, making SQL portability a long-standing challenge in real-world data systems.

Meanwhile, recent advances in large language models (LLMs) have rapidly shifted data interaction from SQL-centric interfaces toward natural-language-centric interfaces. Modern text-to-SQL systems increasingly allow users to express querying intent directly in natural language ~\cite{wang2025mac,pourreza2023din,qin2025route,pourreza2025chase}, reducing the need to manually write dialect-specific SQL queries. This trend raises the possibility of a more universal querying paradigm, where natural language serves as a \textit{unified interface} across heterogeneous database systems. Such a paradigm is increasingly important in the era of data agents and automated analytics, where user requests may involve querying, integrating, and operating over different database systems rather than interacting with a single backend.


However, current text-to-SQL evaluation remains largely SQLite-centric, making it unclear whether existing models can support such cross-system querying in practice.
Existing mainstream benchmarks such as Spider~\cite{yu2018spider} and BIRD~\cite{li2023can} mainly focus on SQLite execution and therefore cannot evaluate executable cross-dialect generalization. 
More recent benchmarks such as Spider 2.0~\cite{lei2025spider} introduce enterprise-level text-to-SQL workflows over different database systems, but their tasks are system-specific rather than aligned realizations of the same natural language intents across dialects. 
As a result, they cannot directly isolate dialect variation from differences in schemas, tasks, or database environments. 
Queries that are executable and semantically correct in one dialect may fail, behave differently, or require substantial reformulation in another database system due to differences in execution semantics, function libraries, ordering behavior, implicit type casting, aggregation rules, and dialect-specific syntax. 
Consequently, current benchmark settings may overestimate the robustness and universality of text-to-SQL systems, while overlooking their sensitivity to dialect variation.


To bridge this gap, we introduce \textsc{UniQL}, a human-verified executable benchmark for cross-dialect text-to-SQL evaluation. 
Built upon the BIRD development set, \textsc{UniQL} aligns 1,534 natural language questions with executable SQL realizations across 16 SQL dialects, yielding 24,544 dialect-specific SQL annotations. 
All dialects share the same natural language intents, aligned database schemas and underlying database contents, enabling controlled evaluation of dialect generalization. 
To construct the benchmark, we develop a hybrid pipeline that combines database migration, tool-based translation, LLM-based translation, self-reflection with execution feedback, iterative translation rule evolution, and human validation for long-tail cases. 
Using \textsc{UniQL}, we evaluate a broad set of open-source and closed-source LLMs in an inference-only setting. 
Our results reveal a gap that is largely hidden by existing SQLite-centered evaluation: current LLMs can often solve an intent in some database systems, but fail to express the same intent consistently across dialects. 
Even strong models solve only about half of the benchmark on average, exhibit large model--dialect interactions, and show limited transfer from SQLite correctness to other database systems. 
These findings suggest that dialect-universal text-to-SQL is not merely a matter of improving overall model capability, but requires explicit evaluation and modeling of cross-dialect robustness.
In summary, the contributions of this paper are as follows:
\begin{itemize}[leftmargin=25pt, itemsep=0pt, topsep=2pt, partopsep=0pt]
    \item We introduce \textsc{UniQL}, a human-verified executable benchmark that aligns the same natural language intents with SQL realizations across 16 dialects.

    \item We propose a SQL translation framework for constructing \textsc{UniQL}, which integrates tool-based translation, LLM-based translation, execution-based verification, self-reflection with execution feedback, iterative translation rule evolution, and human validation.

    \item We comprehensively evaluate open-weight and closed-weight LLMs on \textsc{UniQL}, showing that current models still struggle with dialect-universal text-to-SQL generation.
\end{itemize}

\section{Related Work}

Text-to-SQL benchmarks have been widely studied for evaluating executable SQL generation from natural language. 
Early semantic parsing and natural language interface datasets, such as ATIS~\cite{dahl1994expanding}, GeoQuery~\cite{zelle1996learning}, and Restaurants~\cite{tang2000automated}, focused on domain-specific database querying. 
Subsequent work revisited text-to-SQL evaluation methodology and standardized multiple datasets~\cite{finegan2018improving}. 
Later benchmarks scaled the task from single-table queries in WikiSQL~\cite{zhong2017seq2sql} to complex cross-domain multi-table queries in Spider~\cite{yu2018spider}, and further to value-grounded realistic queries in BIRD~\cite{li2023can}. 
However, these benchmarks are primarily SQLite-based. 
Recent benchmarks such as Spider 2.0~\cite{lei2025spider} cover multiple database systems, but their tasks are system-specific rather than aligned realizations of the same natural language intents across dialects. 
Conversational datasets such as SParC~\cite{yu2019sparc} and CoSQL~\cite{yu2019cosql} focus on multi-turn interactions but also remain largely single-dialect. 
In contrast, \textsc{UniQL} aligns the same natural language intents with executable SQL realizations across 16 human-verified dialects, enabling controlled evaluation of cross-dialect robustness.

Another related line of work studies SQL dialect translation, which rewrites a source SQL query into an equivalent query for another database system. 
Rule-based systems such as SQLGlot~\cite{glot}, JOOQ~\cite{jooq}, and SQLines~\cite{sqlines} rely on manually maintained dialect mappings, while recent LLM-based or hybrid systems such as MALLET~\cite{ngom2024mallet}, RISE~\cite{xie2026rise}, and CrackSQL~\cite{zhou2025cracksql,zhou2025cracking} incorporate LLMs, execution feedback, or dialect-specific rules. 
Recent benchmarks such as PARROT~\cite{zhou2026parrot} further evaluate cross-system SQL-to-SQL translation across many production-grade database systems. 
These efforts are complementary to \textsc{UniQL}: they focus on SQL-to-SQL portability, whereas \textsc{UniQL} evaluates whether models can directly generate executable, dialect-specific SQL from natural language under aligned cross-dialect conditions.
\section{Dataset Construction}
\label{sec:construction}

\begin{figure*}[t!]
    \centering
    \vspace{-3mm}
    \includegraphics[width=\textwidth, trim={0.5cm 2.7cm 0.7cm 0.5cm}]{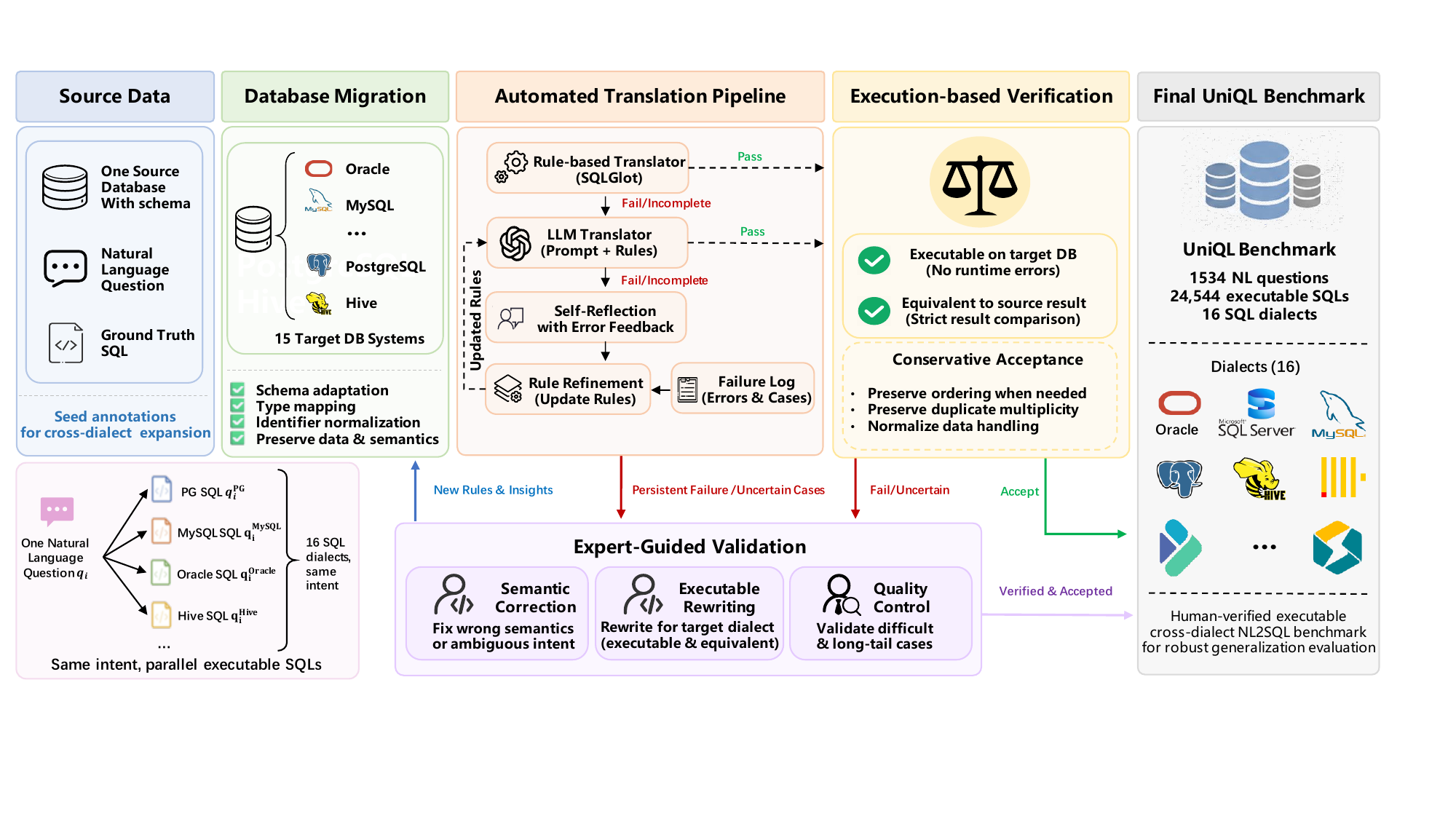}
    \vskip -2mm
    \caption{\textsc{UniQL} construction pipeline. \textsc{UniQL} extends the widely-used BIRD SQLite development set to 16 SQL dialects through database migration, hybrid SQL translation, execution-based verification, iterative translation rule evolution, and human validation. 
   }
    \vspace{-5mm}
    \label{fig:architecture}
\end{figure*}


Constructing a high-quality cross-dialect text-to-SQL benchmark requires more than translating SQL strings. 
A translated query must be executable on the target database system and must preserve the intent of the original natural language question. 
The overall construction pipeline of \textsc{UniQL} is presented in Figure~\ref{fig:architecture}.
\textsc{UniQL} extends the BIRD development set from SQLite to 16 SQL dialects, producing 24,544 executable SQL annotations for 1,534 natural language questions. 
SQLite serves as the source dialect, while the other 15 target dialects are constructed through a hybrid pipeline that combines database migration, automated SQL translation, execution-based verification, iterative translation rule evolution, and human validation.

\subsection{Database Migration}

Since executable cross-dialect evaluation requires running SQL queries on actual database systems, we first migrate the original BIRD databases to each target system. 
Although the migrated databases are designed to preserve the original schemas and values as much as possible, exact one-to-one migration is not always possible across heterogeneous DBMSs. 
Different systems vary in type names, identifier rules, case sensitivity, namespace organization, reserved keywords, and supported constraints. 
For example, some systems organize data through schemas or users rather than standalone database namespaces, while others impose different conventions on table and column names. 
We therefore perform lightweight schema and data normalization during migration, including type mapping, identifier normalization, namespace adaptation, and necessary formatting changes.
This migration step provides the execution environment needed for validating translated SQL queries under each target dialect.


\subsection{Automated Translation and Verification}

Given a NL question \(x\), its source SQLite query \(q_s\), the source database \(D_s\), and the migrated target database \(D_t\), the goal is to construct a target query \(q_t\) in dialect \(\tau_t\) such that \(q_t\) is executable on \(D_t\) and preserves the semantics of \(q_s\). 
We first apply a deterministic rule-based translator (e.g., SQLglot~\cite{glot}) to obtain an initial target query:
\begin{equation}
    q_t^{(0)} = \mathcal{F}_{tool}(q_s, \tau_s, \tau_t),
\end{equation}
where \(\tau_s\) is SQLite and \(\tau_t\) is the target dialect. 
This step efficiently handles common syntactic mappings and standard SQL constructs.

The translated query is then checked through execution-based verification. 
Let \(E(q, D)\) denote the execution result of query \(q\) on database \(D\). 
A translated query is automatically accepted only when it executes successfully on the target database and its result is equivalent to the source execution result under our conservative verification protocol:
\begin{equation}
    \mathcal{V}(q_s, q_t, D_s, D_t) = 
    \mathbb{I}\left[
    E(q_t, D_t) \equiv E(q_s, D_s)
    \right].
\end{equation}

Here, we use a conservative acceptance criterion during construction detailed in Section~\ref{sec:metrics}. 
Standard execution accuracy in prior single-dialect benchmarks often compares query outputs after converting them into unordered sets. 
This can introduce false positives in cross-dialect construction, since ordering information and duplicate multiplicities may be discarded. 
Therefore, our automatic verification preserves ordering when the query has explicit ordering semantics and preserves duplicate-sensitive outputs in unordered comparison. 
At the construction stage, this conservative verification strategy ensures the quality of the automatically accepted translations.

If the rule-based translation cannot be automatically accepted, we invoke an LLM-based translator conditioned on the source SQL, target schema, dialect information, and the current rule set \(\mathcal{R}\):
\begin{equation}
    q_t^{(1)} = 
    \mathcal{F}_{LLM}(q_s, S_t, \tau_t, \mathcal{R}),
\end{equation}
where \(S_t\) denotes the target schema. 
When the generated query still fails equivalence verification, the model enters a bounded self-reflection loop. 
At iteration \(k\), the feedback object \(F_k\) contains the previous target SQL, execution errors, or result mismatches. 
The model then refines the query as:
\begin{equation}
    q_t^{(k+1)} =
    \mathcal{F}_{reflection}(q_s, q_t^{(k)}, F_k, S_t, \tau_t, \mathcal{R}).
\end{equation}
This process repeats until the query is automatically accepted or the maximum number of refinement rounds is reached. In our implementation, the LLM translator is instantiated with GPT-5-mini. 
For each failed translation, we allow up to three execution-feedback reflection rounds before routing the case to later construction stages. 
The translator prompt template is provided in Appendix~\ref{app:prompt}.

\subsection{Iterative Translation Rule Evolution}

Automatic translation failures are not treated as isolated query-level errors. 
Instead, we collect them into a failure log and abstract recurring error patterns into reusable dialect transformation rules. 
For each target dialect, let \(\mathcal{L}_{fail}\) denote the set of failed translations after execution-guided refinement. 
These failures are analyzed together with the current rule set \(\mathcal{R}^n\) and dialect documentation to produce an updated rule set:
\begin{equation}
    \mathcal{R}^{n+1} =
    \mathcal{G}_{rule}(\mathcal{R}^{n}, \mathcal{L}_{fail}, \mathcal{D}_{doc}),
\end{equation}
where \(\mathcal{G}_{rule}\) denotes the rule summarization process and \(\mathcal{D}_{doc}\) denotes target-dialect documentation or tutorials.

The refined rule set is then incorporated into subsequent translation attempts. 
This feedback loop allows the construction pipeline to progressively handle repeated dialect-specific failures, such as function rewriting, type conversion, date and time operations, aggregation behavior, or system-specific syntax constraints. 
In this way, the pipeline combines the efficiency of deterministic translation, the flexibility of LLM-based rewriting, and the reusability of accumulated dialect knowledge.
We instantiate the rule summarizer with Gemini-2.5-Pro and run rule evolution for three rounds. 
The summarizer prompt template is provided in Appendix~\ref{app:prompt}.

\subsection{Human Verification}

After automated translation, self-reflection, and rule refinement, the remaining cases are routed to human verification. 
Human verification serves as the final quality-control stage for examples that cannot be reliably accepted by automatic execution verification. 
These cases include genuine translation failures, unsupported dialect constructs, and examples where source and target executions differ due to dialect-dependent behavior or under-specified semantics in the original SQL. 
For example, different systems may return different row orders when the query does not fully specify tie-breaking, or duplicate-sensitive outputs may require semantic judgment beyond simple set comparison.

Given a candidate target query \(q_t\), the original NL question \(x\), and the target database \(D_t\), human annotators correct or rewrite the query while preserving the intent of the NL question:
\begin{equation}
    q_t^{*} = \mathcal{A}(x, q_s, q_t, D_t, \tau_t),
\end{equation}
where \(q_t^{*}\) is the final human-verified SQL annotation for the target dialect. 
Each routed case is independently reviewed by two annotators, who check both executability and semantic consistency against the original NL question and target database. 
If the candidate SQL is invalid, ambiguous, or not semantically equivalent, annotators rewrite it into an executable target-dialect query. 
Disagreements are resolved through discussion, and the final annotation is obtained only after consensus is reached.

Overall, \textsc{UniQL}'s construction pipeline follows a strict automatic acceptance and human-in-the-loop verification strategy. 
The automatic stages efficiently resolve the majority of translations, while human verification handles long-tail cases that require semantic judgment. 
This process yields complete executable SQL coverage for 15 target dialects and, together with the original SQLite annotations, forms a 16-dialect benchmark for controlled cross-dialect text-to-SQL evaluation.
\section{Data Statistics}

\paragraph{Overall Statistics.}
\textsc{UniQL} is constructed from the BIRD development set~\cite{li2023can} and contains 1,534 natural language questions over 11 databases. 
For each question, we provide executable SQL annotations across 16 SQL dialects, yielding 24,544 dialect-specific SQL queries in total. 
The covered dialects are SQLite, ClickHouse, Doris, Drill, Druid, DuckDB, Hive, MySQL, Oracle, PostgreSQL, Presto, Spark, StarRocks, Teradata, Trino, and T-SQL. 
SQLite serves as the source dialect, while the remaining 15 dialects are constructed through our hybrid construction pipeline.
Since all dialects share the same natural language questions, aligned schemas and database contents, \textsc{UniQL} enables controlled evaluation of dialect generalization without conflating dialect variation with changes in tasks, schemas, or domains.

\paragraph{Comparison with Existing Benchmarks.}
Table~\ref{tab:benchmark_comparison} compares \textsc{UniQL} with representative text-to-SQL benchmarks. 
Existing benchmarks are either single-dialect, such as WikiSQL~\cite{zhong2017seq2sql}, Spider 1.0~\cite{yu2018spider}, and BIRD~\cite{li2023can}, or cover multiple systems without aligning the same natural language intents across dialects, such as Spider 2.0-lite~\cite{lei2025spider}. 
In contrast, \textsc{UniQL} expands each natural language question into executable SQL annotations across 16 dialects, making it suitable for controlled cross-dialect text-to-SQL evaluation.

\begin{table}[t]
\centering
\footnotesize
\setlength{\tabcolsep}{2.5pt}
\renewcommand{\arraystretch}{1.10}
\caption{Comparison between \textsc{UniQL} and representative text-to-SQL benchmarks. }
\label{tab:benchmark_comparison}
\vspace{-2mm}
\begin{tabular}{ccccc@{}}
\toprule
\textbf{Benchmark} 
& \textbf{\#Q} 
& \textbf{\#SQL} 
& \textbf{\#Dialect}
& \textbf{Aligned SQL}\\
\midrule
WikiSQL
& 80,654 & 77,840 & 1 
& --  \\

BIRD
& 12,751 & 12,751 & 1 
& -- \\

Spider 1.0
& 10,181 & 5,693 & 1 
& -- \\

Spider 2.0-lite
& 547 & 547 & 6 
& \xmark \\
\midrule
\textsc{UniQL} 
& 1,534 & 24,544 & 16 
& \cmark  \\
\bottomrule
\end{tabular}
\vspace{-3mm}
\end{table}

\begin{figure}[t!]
    \centering
    \vspace{-2mm}
    \includegraphics[width=0.48\textwidth]{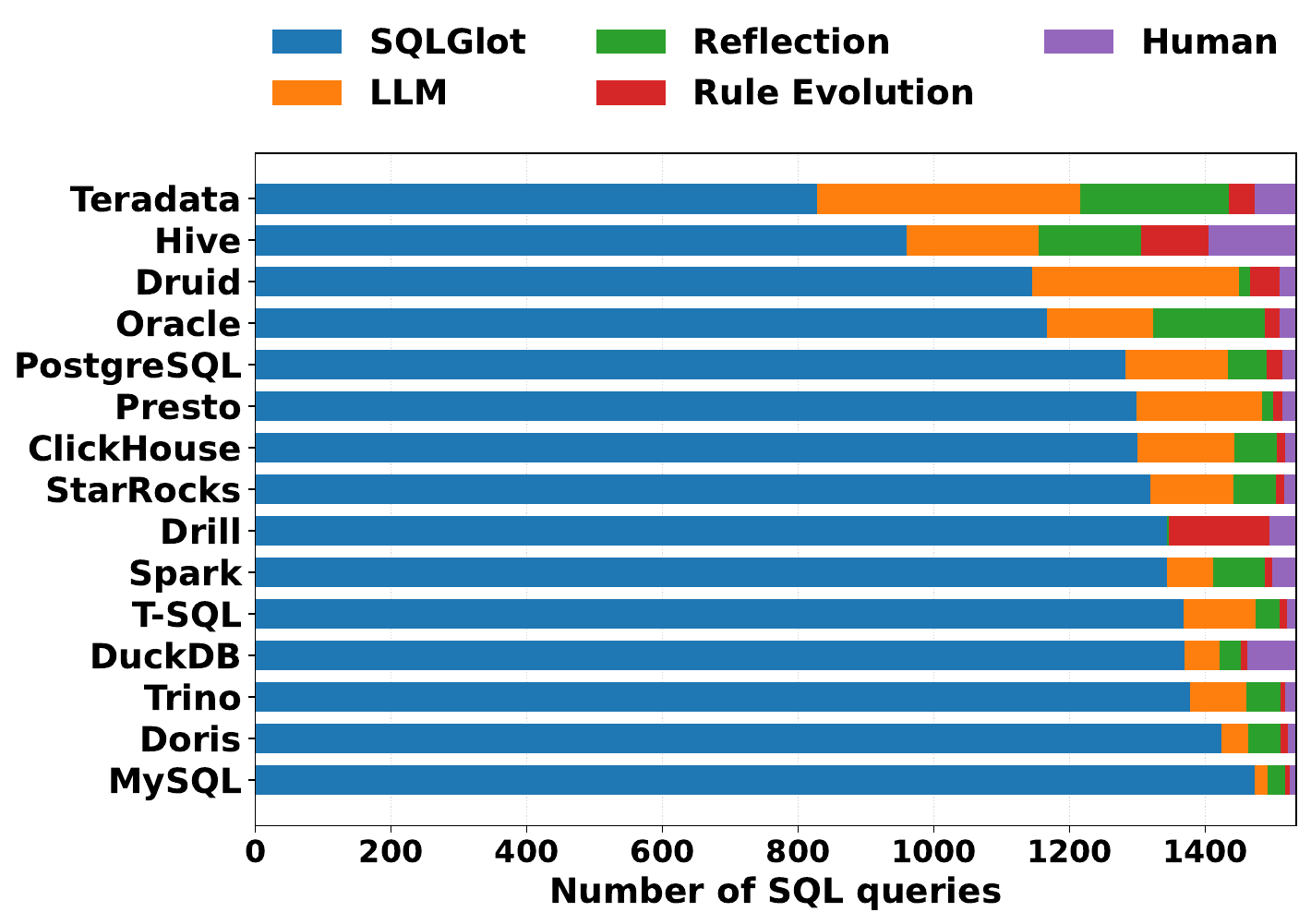}
    \vskip -2.5mm
    \caption{Construction pipeline statistics for the 15 target dialects in \textsc{UniQL}.}
    \vspace{-5mm}
    \label{fig:statistics}
\end{figure}

\paragraph{Construction Pipeline Statistics.}
Figure~\ref{fig:statistics} summarizes how the 15 target dialects are completed through different stages of the construction pipeline. 
All target dialects are eventually completed with 1,534 executable SQL annotations, ensuring full alignment with the original BIRD development questions. 
The statistics reveal substantial variation in construction difficulty across dialects: several dialects, such as MySQL and Doris, can largely be handled by SQLGlot, while Teradata and Hive require more LLM translation, reflection, rule refinement, or human annotation. 
This variation shows that reliable cross-dialect benchmark construction cannot be reduced to one-shot SQL translation, since different database systems expose different long-tail incompatibilities in syntax, functions, type handling, and execution behavior.

\section{Evaluation Metrics}
\label{sec:metrics}

We use execution accuracy (EX) as the main evaluation metric in \textsc{UniQL}. 
The same execution-based protocol is used for both automatic verification during dataset construction and model evaluation during experiments. 
During construction, EX serves as a conservative acceptance criterion: a translated SQL query is automatically accepted only when it executes successfully on the target database and produces results consistent with the source query; otherwise, uncertain cases are routed to human verification. 
During model evaluation, the predicted SQL and the gold SQL are executed on the same target database, and the prediction is considered correct only when their execution results match.

Different from BIRD-style EX implementations that convert query outputs into unordered sets, our protocol preserves ordering when explicit ordering semantics are present and preserves duplicate multiplicities for unordered outputs. 
This stricter comparison avoids accepting predictions that return the correct rows in the wrong order or with incorrect duplicate counts. 
We provide the detailed formal definition of the metric in Appendix~\ref{app:metric}.
\section{Experiments}

\begin{table*}[t]
\centering
\vskip -2mm
\setlength{\tabcolsep}{2.0pt}
\renewcommand{\arraystretch}{1.08}
\caption{Main results on \textsc{UniQL}. We report execution accuracy (\%) across 16 SQL dialects. All models are evaluated in an inference-only setting without task-specific training. DS means DeepSeek.}
\vskip -2mm
\label{tab:main_results}
\resizebox{\textwidth}{!}{
\begin{tabular}{lccccccccccccccccc}
\toprule
\multirow{2}{*}{\textbf{Model}} &
\multicolumn{16}{c}{\textbf{SQL Dialect}} & \multirow{2}{*}{\textbf{Avg.}} \\
\cmidrule(lr){2-17}
& \rotatebox{50}{\textbf{ClickHouse}}
& \rotatebox{50}{\textbf{Doris}}
& \rotatebox{50}{\textbf{Drill}}
& \rotatebox{50}{\textbf{Druid}}
& \rotatebox{50}{\textbf{DuckDB}}
& \rotatebox{50}{\textbf{Hive}}
& \rotatebox{50}{\textbf{MySQL}}
& \rotatebox{50}{\textbf{Oracle}}
& \rotatebox{50}{\textbf{PostgreSQL}}
& \rotatebox{50}{\textbf{Presto}}
& \rotatebox{50}{\textbf{Spark}}
& \rotatebox{50}{\textbf{SQLite}}
& \rotatebox{50}{\textbf{StarRocks}}
& \rotatebox{50}{\textbf{T-SQL}}
& \rotatebox{50}{\textbf{Teradata}}
& \rotatebox{50}{\textbf{Trino}} \\

\midrule
\multicolumn{17}{l}{\textit{Closed-Source LLMs}} \\
GPT-3.5-Turbo & 35.14 & 38.33 & 29.20 & 22.75 & 37.09 & 36.18 & 40.81 & 49.35 & 36.44 & 36.31 & 39.37 & 41.13 & 39.18 & 37.74 & 23.86 & 35.85 & 36.17 \\
GPT-5-mini & 51.56 & 49.80 & 48.83 & 37.74 & 52.74 & 54.04 & 54.30 & 60.56 & 51.37 & 50.98 & 52.93 & 52.48 & 51.89 & 53.85 & 35.85 & 49.02 & 50.50 \\
GPT-5.1-codex & 52.61 & 49.93 & 48.11 & 38.20 & 52.93 & 55.48 & 54.89 & 49.67 & 52.35 & 52.41 & 53.65 & 53.78 & 52.41 & 54.63 & 35.07 & 50.72 & 50.43 \\
Gemini-2.5-Pro & 53.98 & 51.76 & 50.85 & 37.42 & 56.19 & 59.32 & 57.82 & 34.29 & 53.59 & 57.89 & 56.65 & 59.78 & 54.43 & 55.61 & 38.40 & 55.61 & 52.10 \\
Claude-4.5-Sonnet & 56.84 & 52.74 & 52.28 & 39.90 & 55.28 & 59.58 & 58.54 & 63.75 & 54.95 & 58.08 & 56.13 & 59.84 & 56.06 & 56.26 & 37.74 & 56.06 & 54.63 \\
\midrule
\multicolumn{17}{l}{\textit{Open-Source LLMs}} \\
Qwen3-1.7B & 33.90 & 35.07 & 27.57 & 18.71 & 35.27 & 36.70 & 35.85 & 43.48 & 32.92 & 32.46 & 35.27 & 35.40 & 36.31 & 32.66 & 22.43 & 34.16 & 33.01 \\
Qwen3-4B & 43.02 & 44.52 & 41.72 & 26.53 & 44.26 & 46.74 & 46.61 & 44.92 & 42.70 & 41.72 & 43.94 & 46.94 & 45.50 & 46.41 & 29.60 & 41.46 & 42.29 \\
Qwen3-8B & 46.54 & 46.15 & 44.13 & 31.29 & 47.00 & 44.85 & 50.20 & 51.11 & 46.35 & 32.99 & 47.46 & 49.09 & 47.46 & 48.57 & 29.34 & 44.20 & 44.17 \\
Qwen3-32B & 49.74 & 47.20 & 46.54 & 33.83 & 49.93 & 51.96 & 52.02 & 53.06 & 48.50 & 47.78 & 49.22 & 53.39 & 50.52 & 51.69 & 32.27 & 46.61 & 47.77 \\
Llama-3-8B-Inst & 20.80 & 23.99 & 18.45 & 16.56 & 22.29 & 22.23 & 22.03 & 31.23 & 22.43 & 22.23 & 23.34 & 23.60 & 24.12 & 21.12 & 15.45 & 22.75 & 22.04 \\
Llama-3-70B-Inst & 40.16 & 40.09 & 37.74 & 26.92 & 41.72 & 42.63 & 43.74 & 51.56 & 40.16 & 38.01 & 42.89 & 42.24 & 42.57 & 40.61 & 26.47 & 39.05 & 39.78 \\
DeepSeek-Coder-16B & 32.46 & 31.10 & 32.07 & 21.38 & 32.72 & 34.88 & 36.57 & 45.44 & 31.88 & 31.55 & 34.42 & 35.01 & 34.88 & 31.68 & 19.95 & 32.53 & 32.41 \\
DeepSeek-v4-flash & 48.04 & 46.94 & 46.28 & 34.22 & 49.02 & 52.80 & 52.54 & 57.50 & 48.50 & 49.02 & 51.56 & 53.46 & 50.20 & 51.43 & 30.90 & 47.59 & 48.12 \\
\midrule
\textbf{Avg.} & 43.45 & 42.89 & 40.29 & 29.65 & 44.34 & 45.95 & 46.61 & 48.92 & 43.24 & 42.42 & 45.14 & 46.63 & 45.04 & 44.79 & 29.03 & 42.74 & --\\
\bottomrule
\end{tabular}
}
\vspace{-4mm}
\end{table*}

\subsection{Experimental Setup}
\label{sec:exp_setup}

\paragraph{Evaluation Setting.}
We evaluate all models on the 16 SQL dialects in \textsc{UniQL}. 
For each test instance, the model is given the target SQL dialect, the database schema, and the natural language question, and is required to generate one executable SQL query in the specified dialect. 
All models are evaluated once in an inference-only setting, without task-specific supervised fine-tuning, reinforcement learning, few-shot example selection, or dialect-specific adaptation. 
This setting is designed to measure the out-of-the-box cross-dialect text-to-SQL capability of foundation models.

To ensure a fair comparison, all models use the same prompt template. 
The template is instantiated with different target dialects, schemas, and natural language questions for different test cases, while the instruction format remains unchanged across models. 
The full prompt template is provided in Appendix~\ref{app:prompt}.
For evaluation, we use the same EX protocol as described in Section~\ref{sec:metrics}; although this protocol is stricter than the set-based EX used in some prior benchmarks, it is more appropriate for comparing model predictions against golden SQL executions on the target database, where ordering semantics and duplicate-sensitive outputs should be preserved when relevant.

\paragraph{Models.}
We evaluate both open-source and closed-source LLMs. 
The closed-source models include GPT-3.5-Turbo~\cite{openai2023gpt35turbo}, GPT-5-mini~\cite{singh2026openai}, GPT-5.1-codex~\cite{openai2025gpt51codex}, Gemini-2.5-Pro~\cite{comanici2025gemini}, and Claude-4.5-Sonnet~\cite{claude}. 
The open-source models include the Qwen3 series~\cite{yang2025qwen3}, Llama-3-Instruct models~\cite{dubey2024llama}, DeepSeek-Coder-V2-Lite-16B~\cite{zhu2024deepseekcoderv2}, and DeepSeek-v4-flash~\cite{deepseek2026v4}. 
The Qwen3 models cover multiple parameter scales, allowing us to analyze scaling behavior. 
DeepSeek-Coder-V2-Lite-16B is denoted as DeepSeek-Coder-16B for compactness.

\subsection{Main Results}
\label{sec:main_results}

Table~\ref{tab:main_results} reports execution accuracy across all 16 SQL dialects. 
Overall, \textsc{UniQL} remains challenging for current LLMs. 
The best-performing model, Claude-4.5-Sonnet, achieves an average EX of 54.63\%, followed by Gemini-2.5-Pro with 52.10\% and GPT-5-mini with 50.50\%. 
Even these strong models fail to solve a large fraction of examples, indicating that executable cross-dialect text-to-SQL generation remains far from saturated.

\vspace{0.5mm}
\noindent\textit{Observation 1: Stronger models perform better, but the task remains challenging.}
Frontier API-based models achieve the strongest overall results, with Claude-4.5-Sonnet leading on most dialects. 
However, the best average EX is still only 54.63\%, showing that \textsc{UniQL} exposes substantial remaining limitations even for advanced LLMs. 
Among open-source models, DeepSeek-v4-flash and Qwen3-32B obtain the strongest averages, but they still leave considerable room for improvement.

\vspace{0.5mm}
\noindent\textit{Observation 2: SQLite-only evaluation hides dialect-specific weaknesses.}
Because \textsc{UniQL} aligns the same natural language intents across dialects, it allows direct comparison of model behavior under different target SQL dialects. 
The results show that performance on SQLite does not fully reflect cross-dialect ability. 
For example, Claude-4.5-Sonnet achieves 59.84\% on SQLite but 63.75\% on Oracle and 37.74\% on Teradata. 
These gaps demonstrate that single-dialect benchmarks can miss important model--dialect interactions.

\vspace{0.5mm}
\noindent\textit{Observation 3: Scaling helps within model families.}
The Qwen3 series exhibits a clear scaling trend. 
Average EX increases from 33.01\% for Qwen3-1.7B to 42.29\% for Qwen3-4B, 44.17\% for Qwen3-8B, and 47.77\% for Qwen3-32B. 
A similar trend appears in the Llama family, where Llama-3-70B-Instruct substantially improves over Llama-3-8B-Instruct. 
These results suggest that model capacity benefits cross-dialect SQL generation. 
At the same time, scaling mainly improves overall capability rather than eliminating dialect-level variation: difficult dialects such as Druid, Presto, and Teradata remain challenging even for larger models.

\vspace{0.5mm}
\noindent\textit{Observation 4: Dialect performance reflects model familiarity as well as execution behavior.}
The bottom row of Table~\ref{tab:main_results} shows large differences across target dialects. 
Oracle obtains the highest average EX of 48.92\%, followed by SQLite, MySQL, Hive, Spark, and StarRocks, while Teradata and Druid are the most difficult with 29.03\% and 29.65\%, respectively. 
This pattern should not be interpreted simply as a ranking of intrinsic dialect complexity. 
For instance, Oracle is not easy to translate from SQLite in our construction pipeline, but models still perform strongly on it, likely because Oracle SQL has abundant public documentation, examples, and training exposure. 
Conversely, dialects such as Teradata and Druid may be less represented in model pretraining corpora and involve more system-specific functions, type handling, temporal operations, or execution behavior. 
Thus, \textsc{UniQL} reveals not only syntactic differences among SQL dialects, but also the uneven familiarity and robustness of current LLMs across database ecosystems.

Overall, the results show that \textsc{UniQL} is both challenging and diagnostic: it reveals that current models benefit from stronger general capability, but still lack robust dialect-universal text-to-SQL generation. 
The substantial variation across dialects further supports the need for aligned cross-dialect benchmarks beyond SQLite-centered evaluation.

\subsection{Stratified Performance Analysis}
\label{sec:stratified_analysis}

Beyond dialect-level results, we further analyze model performance under two benchmark partitions: the BIRD question difficulty and the construction source introduced by \textsc{UniQL}. 
These partitions help distinguish whether models fail mainly because the original text-to-SQL problem is complex, or because the target SQL realization belongs to a harder cross-dialect construction path.

\begin{figure}[t!]
    \centering
    \includegraphics[width=0.47\textwidth]{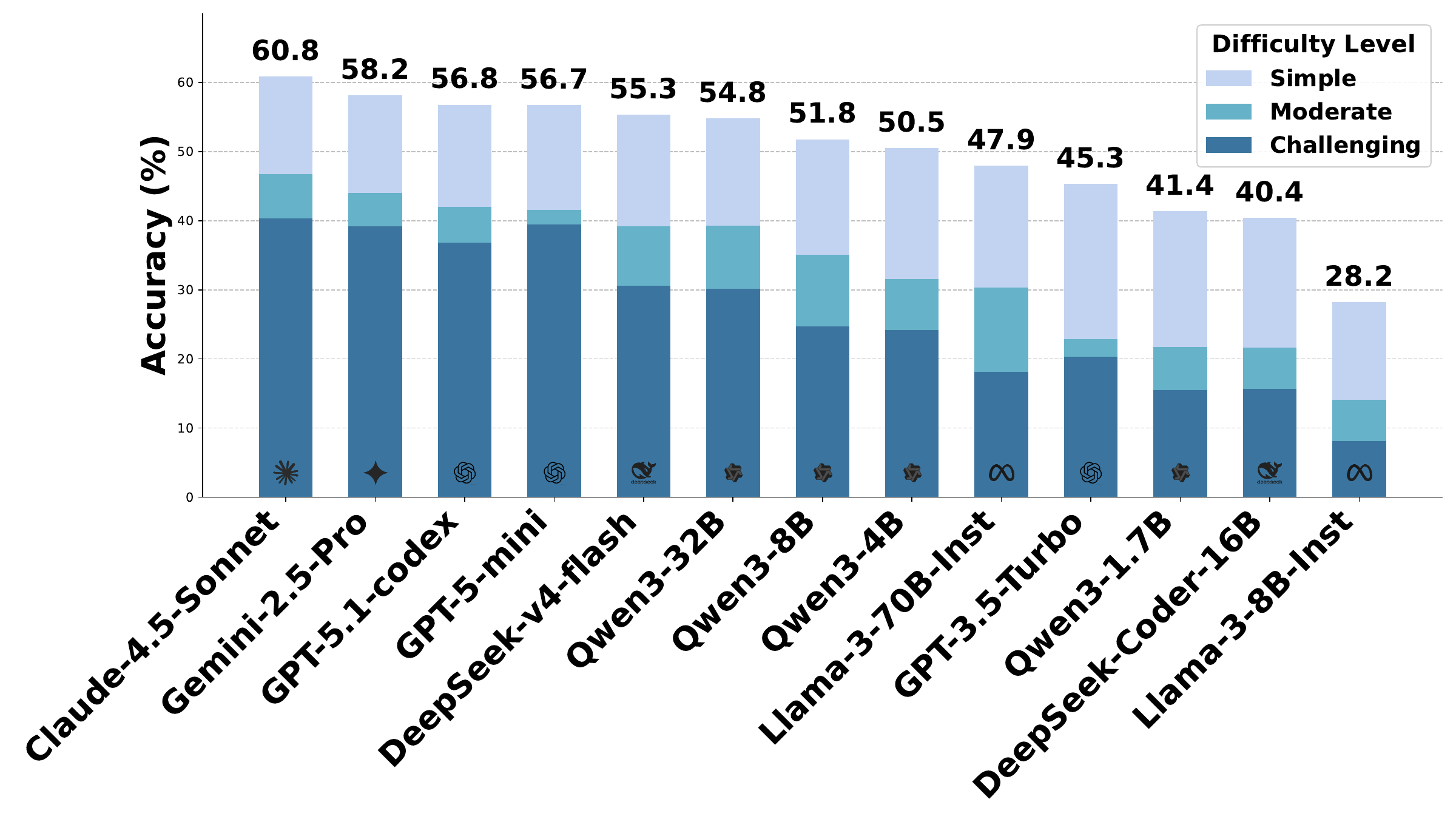}
    \vspace{-2mm}
    \caption{Stratified execution accuracy (\%) of the 13 models by the original BIRD question difficulty. Values are macro-averaged over 16 SQL dialects on UniQL.}
    \label{fig:bird_difficulty}
    \vspace{-3mm}
\end{figure}

\paragraph{Performance by Question Difficulty.}
Figure~\ref{fig:bird_difficulty} reports execution accuracy grouped by the original BIRD difficulty labels. 
Across almost all models, accuracy decreases consistently from simple to moderate and challenging questions. 
For example, Claude-4.5-Sonnet drops from 60.8\% on simple questions to 46.7\% on moderate questions and 40.3\% on challenging questions, while Qwen3-32B drops from 54.8\% to 39.2\% and 30.2\%. 
This trend shows that \textsc{UniQL} preserves the semantic difficulty structure of the original text-to-SQL task: cross-dialect generation is not only a matter of producing dialect-specific syntax, but also requires solving increasingly complex query semantics.


\begin{figure}[t]
    \centering
    \includegraphics[width=\columnwidth, trim={0, 0.8cm, 0, 0}, clip]{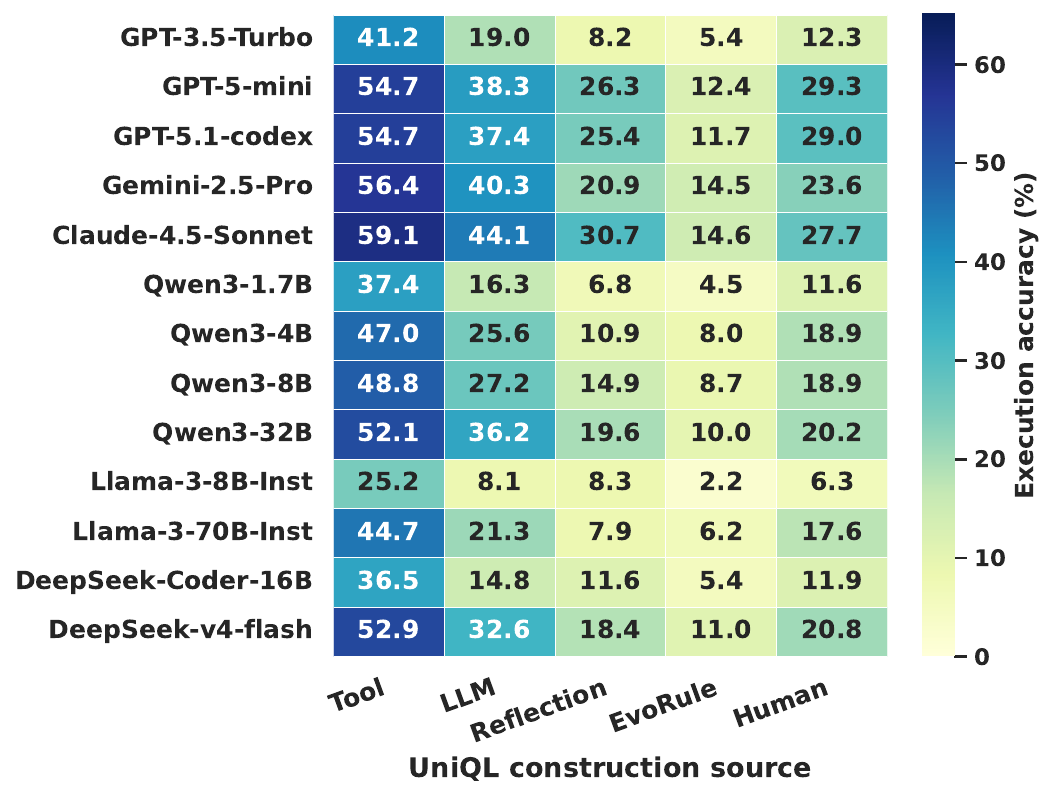}
    \vskip -3mm
    \caption{Execution accuracy by \textsc{UniQL} construction source. Values are macro-averaged over target dialects. }
    \label{fig:construction_source}
    \vspace{-3mm}
\end{figure}

\paragraph{Performance by Construction Source.}
Figure~\ref{fig:construction_source} analyzes performance according to the stage from which the reference SQL annotation is obtained during \textsc{UniQL} construction. 
A clear trend emerges: examples directly resolved by the rule-based tool are substantially easier, while those requiring LLM translation, self-reflection, rule evolution, or human validation are generally more difficult. 
For instance, Claude-4.5-Sonnet achieves 59.1\% on the Tool subset, but only 44.1\%, 30.7\%, and 14.6\% on the LLM, Reflection, and Rule subsets, respectively. 
Similar patterns appear for Qwen3-32B and DeepSeek-v4-flash. 
This indicates that the construction source provides an additional diagnostic signal: examples that require later construction stages often contain more dialect-specific or long-tail behaviors, which are also harder for generation models.

The Human subset does not always have the lowest accuracy. 
This is expected because human verification in \textsc{UniQL} is not a pure difficulty label. 
Some examples enter the human stage because automatic verification cannot safely certify equivalence under database-specific execution behavior, such as under-specified ordering, duplicate-sensitive outputs, or target-engine differences, even when the underlying semantic intent is clear. 
Therefore, the construction-source split should be interpreted as a long-tail and verification-complexity indicator rather than a strict difficulty scale.

\subsection{Cross-Dialect Consistency}

The main results report accuracy by aggregating questions within each dialect. 
However, \textsc{UniQL} is aligned by construction: each question is realized as executable SQL across 16 dialects. 
This allows us to evaluate a stricter property, which we call \emph{cross-dialect consistency}: for the same intent, how many dialect realizations can a model answer correctly? 
For each question, we count the number of dialects answered correctly, ranging from 0 to 16. 
We also report SQLite$\rightarrow$15, which measures the percentage of SQLite-correct questions that remain correct across other 15 dialects.

Table~\ref{tab:consistency_summary} shows that high average execution accuracy does not imply dialect-universal correctness. 
For example, Claude-4.5-Sonnet answers 8.74 dialect realizations correctly per question on average, but only 20.14\% of questions are answered correctly in all 16 dialects. 
Similarly, Gemini-2.5-Pro has a high mean correct count of 8.34, yet its All-16 consistency is only 9.19\%. 
These results indicate that many questions are solved only partially across dialects: a model may correctly generate SQL for some database systems while failing to express the same intent in others.

The SQLite$\rightarrow$15 metric further shows that SQLite correctness is not a reliable proxy for cross-dialect robustness. 
Even among SQLite-correct questions, only 33.66\% for Claude-4.5-Sonnet and 34.29\% for GPT-5-mini remain correct across all other dialects. 
For Gemini-2.5-Pro, this ratio is only 15.38\%, despite its strong average performance. 
Thus, solving the SQLite realization of an intent does not guarantee that the same intent can be realized correctly in other SQL dialects.

\begin{table}[t]
\centering
\small
\setlength{\tabcolsep}{4.0pt}
\renewcommand{\arraystretch}{1.06}
\caption{Cross-dialect consistency on \textsc{UniQL}. 
Mean reports the average correct dialects per question.}
\label{tab:consistency_summary}
\resizebox{\columnwidth}{!}{
\begin{tabular}{lrrr}
\toprule
\textbf{Model} 
& \textbf{Mean} 
& \textbf{All-16} 
& \textbf{SQLite$\rightarrow$15} \\
\midrule
\multicolumn{4}{l}{\textit{Closed-Source LLMs}} \\
GPT-3.5-Turbo      & 5.79 & 9.13  & 22.19 \\
GPT-5-mini         & 8.08 & 17.99 & 34.29 \\
GPT-5.1-codex      & 8.07 & 14.34 & 26.67 \\
Gemini-2.5-Pro     & 8.34 & 9.19  & 15.38 \\
Claude-4.5-Sonnet  & 8.74 & 20.14 & 33.66 \\
\midrule
\multicolumn{4}{l}{\textit{Open-Source LLMs}} \\
Qwen3-1.7B         & 5.28 & 5.80  & 16.39 \\
Qwen3-4B           & 6.77 & 8.80  & 18.75 \\
Qwen3-8B           & 7.17 & 12.06 & 24.57 \\
Qwen3-32B          & 7.64 & 14.41 & 26.98 \\
Llama-3-8B-Inst    & 3.53 & 2.54  & 10.77 \\
Llama-3-70B-Inst   & 6.37 & 14.08 & 33.33 \\
DeepSeek-Coder-16B & 5.00 & 4.63  & 13.22 \\
DeepSeek-v4-flash  & 7.70 & 13.89 & 25.98 \\
\bottomrule
\end{tabular}
}
\end{table}

Overall, the consistency results strengthen the main observation that single-dialect evaluation is insufficient. 
Average EX measures performance after aggregating over dialects, while All-16 consistency asks whether a model can preserve correctness across all dialect realizations. 
The large gap between these two views shows that current models are still far from serving as dialect-universal natural language database interfaces.

\section{Conclusion}
\label{sec:conclusion}

We introduced \textsc{UniQL}, a human-verified benchmark for evaluating cross-dialect text-to-SQL generation. \textsc{UniQL} aligns 1,534 natural language intents with executable SQL realizations across 16 SQL dialects, yielding 24,544 dialect-specific annotations. Through database migration, hybrid translation, execution-guided verification, iterative rule summarization, and human validation, \textsc{UniQL} enables controlled evaluation of whether models can preserve the same query intent across different database systems. Our experiments show that current LLMs remain far from dialect-universal: even strong models solve only around half of the benchmark on average, exhibit substantial variation across dialects, and often fail to transfer SQLite success to other database systems. These findings highlight the need for aligned cross-dialect benchmarks and dialect-aware text-to-SQL methods.

\section{Limitations and Future Work}
\label{sec:limitations}

\textsc{UniQL} prioritizes controlled cross-dialect alignment: the same natural language intents, aligned schemas and database contents are preserved across 16 SQL dialects. 
This design enables direct comparison across dialects, but it also limits the use of target-system-specific data models. 
Our database migration process largely preserves the original BIRD schemas, which are SQLite-derived and mostly flat relational tables. 
Therefore, the current benchmark does not systematically cover native features such as PostgreSQL or BigQuery JSON querying, Spark SQL and Hive complex types such as arrays, maps, and structs, or analytical-system features in ClickHouse and Druid for large-scale aggregation and time-series analysis.

Similarly, the SQL annotations in \textsc{UniQL} are designed to preserve the original query intent rather than to maximize dialect-specific idioms. 
A target SQL query may therefore use a portable formulation even when the target system provides a more native expression. 
Future work can extend \textsc{UniQL} by redesigning target databases with native data models and by creating natural language questions that explicitly require dialect-specific capabilities, such as JSON operators, nested data access, array functions, partition-aware queries, or time-granularity operations. 
Such extensions would complement the current benchmark by evaluating not only cross-dialect portability, but also dialect-native text-to-SQL generation in real-world heterogeneous database environments.

Another limitation is that execution-based verification, even with our stricter protocol, cannot fully guarantee semantic equivalence between translated SQL queries and the original SQL. 
Two queries may return the same result on the current database instance while encoding different logical conditions, especially when the database contents do not expose the difference. 
For example, different predicates, joins, or aggregation conditions may be indistinguishable on the observed data but diverge under other database instances. 
\textsc{UniQL} mitigates this risk through conservative automatic acceptance, execution-guided checking, and human verification for ambiguous or long-tail cases. 
Nevertheless, execution agreement should be understood as strong empirical evidence of equivalence on the benchmark databases rather than a complete formal guarantee of semantic equivalence.

\bibliography{custom}

\clearpage
\appendix


\section{Detailed Execution Accuracy Protocol}
\label{app:metric}

We use execution accuracy (EX) as the main evaluation metric in \textsc{UniQL}. 
The same execution-based protocol is used in two stages: automatic verification during dataset construction and model evaluation during experiments. 
During construction, EX is used as a conservative acceptance criterion for automatically translated SQL queries. 
This criterion may reject some translations that are semantically reasonable but differ in execution results due to dialect-dependent database behavior. 
This is intentional: automatic acceptance favors precision over recall, and uncertain cases are routed to human verification. 
During model evaluation, the same protocol is appropriate because the predicted SQL and the gold SQL are executed on the same target database system. 
In this setting, differences in ordering or duplicate multiplicity should be preserved rather than ignored.

Given a predicted SQL query \(\hat{q}\), the gold SQL query \(q^*\), and the database \(B\), we first execute both queries on the same database:
\begin{equation}
    \hat{R} = E(\hat{q}, B), \qquad R^* = E(q^*, B),
\end{equation}
where \(E(q,B)\) denotes the execution result of query \(q\) on \(B\). 
If \(\hat{q}\) fails to execute, the prediction is marked as incorrect. 
Otherwise, we compare \(\hat{R}\) and \(R^*\) according to the query semantics.

Unlike BIRD-style EX implementations that convert query outputs into unordered sets, our metric preserves ordering and duplicate multiplicities when needed. 
For queries with explicit ordering semantics, such as those containing \texttt{ORDER BY}, we compare the outputs as ordered lists:
\begin{equation}
    \mathrm{EX}(\hat{q}, q^*) = \mathbb{I}[\hat{R} = R^*].
\end{equation}
This prevents a prediction with the correct rows but wrong order from being incorrectly accepted.

For unordered queries, result order is ignored, but duplicate multiplicity is still preserved. 
Instead of converting outputs into sets, we use multiset-style matching: each row in \(\hat{R}\) must be matched with and removed from \(R^*\), and the prediction is correct only if all rows can be matched exactly and no unmatched rows remain. 
Formally, for unordered outputs we require:
\begin{equation}
    \mathrm{MultiSet}(\hat{R}) = \mathrm{MultiSet}(R^*).
\end{equation}
This avoids treating outputs with different numbers of duplicate rows as equivalent.

Overall, this protocol provides stricter and more faithful executable evaluation for cross-dialect text-to-SQL than set-based result comparison.

\section{Prompt Templates}
\label{app:prompt}

We provide the prompt templates used in \textsc{UniQL} for both model inference and benchmark construction. 
The inference prompt is used to evaluate text-to-SQL models across different SQL dialects. 
The translator prompt is used to convert SQL queries from a source dialect to a target dialect, while the rule summarizer prompt is used to derive reusable dialect-specific translation rules from failed translation cases. 
For each instance, placeholders such as \texttt{\{dialect\}}, \texttt{\{db\_details\}}, \texttt{\{question\}}, \texttt{\{evidence\}}, \texttt{\{schema\}}, and \texttt{\{error\_logs\}} are instantiated with the corresponding database, query, and execution feedback information.

\vspace{0.5em}
\Needspace{12\baselineskip}
\begin{tcolorbox}[
    enhanced,
    breakable,
    colback=blue!2,
    colframe=blue!50!black,
    title=\textbf{Prompt Template for Text-to-SQL Inference},
    fonttitle=\bfseries,
    arc=1.0mm,
    boxrule=0.5pt,
    left=1.0mm,
    right=1.0mm,
    top=0.8mm,
    bottom=0.8mm,
    before skip=4pt,
    after skip=6pt,
    width=\linewidth
]
\scriptsize
\begin{Verbatim}[breaklines=true, breaksymbolleft={}, fontsize=\scriptsize]
Task Overview:
You are a data science expert. Below, you are provided with a database schema and a natural language question. Your task is to understand the schema and generate a valid SQL query to answer the question.

Database system:
{dialect}

Database Schema:
{db_details}
This schema describes the database's structure, including tables, columns, primary keys, foreign keys, and any relevant relationships or constraints.

Question:
{question}

Evidence:
{evidence}

Instructions:
- Make sure you only output the information that is asked in the question. If the question asks for a specific column, make sure to only include that column in the SELECT clause, nothing more.
- The generated query should return all of the information asked in the question without any missing or extra information.
- Before generating the final SQL query, please think through the steps of how to write the query.

Output Format:
In your answer, please enclose the generated SQL query in a code block:
```sql
-- Your SQL query
```
Take a deep breath and think step by step to find the correct SQL query.
\end{Verbatim}
\end{tcolorbox}

\Needspace{12\baselineskip}
\begin{tcolorbox}[
enhanced,
breakable,
colback=green!2,
colframe=green!45!black,
title=\textbf{Prompt Template for the LLM Translator},
fonttitle=\bfseries,
arc=1.0mm,
boxrule=0.5pt,
left=1.0mm,
right=1.0mm,
top=0.8mm,
bottom=0.8mm,
before skip=4pt,
after skip=6pt,
width=\linewidth
]
\scriptsize
\begin{Verbatim}[breaklines=true, breaksymbolleft={}, fontsize=\scriptsize]
You are an expert SQL translator. Your task is to convert a {source_dialect} query into an executable {target_dialect} query.

Target {target_dialect} Schema (Reference for Exact Table/Column Names and Types):
{schema}

Source {source_dialect} SQL:
{source_sql}

General Translation Rules:

Schema Fidelity: The provided Schema is the ABSOLUTE TRUTH. Do not invent columns or tables.
Logic Preservation: Preserve the logic of the source SQL (filters, joins, ordering) as much as possible, adapting syntax to the target dialect.
Data Types: Be careful with Date/Time and String/Number comparisons.

Specific Dialect Rules (Iteratively Refined):
{specific_rules}

Output Format:
In your answer, please enclose the translated SQL query in a code block:
```sql
-- Your SQL query
```

\end{Verbatim}
\end{tcolorbox}

\begin{tcolorbox}[
enhanced,
breakable,
colback=orange!3,
colframe=orange!60!black,
title=\textbf{Prompt Template for the Rule Summarizer},
fonttitle=\bfseries,
arc=1.0mm,
boxrule=0.5pt,
left=1.0mm,
right=1.0mm,
top=0.8mm,
bottom=0.8mm,
before skip=4pt,
after skip=6pt,
width=\linewidth
]
\scriptsize
\begin{Verbatim}[breaklines=true, breaksymbolleft={}, fontsize=\scriptsize]
You are an expert Database Administrator and SQL Architect.
We are translating SQL queries from {source_dialect} to {target_dialect}.
We have a list of failed translations where the translated SQL result did not match the ground truth.

Your task is to analyze these error logs and summarize specific, actionable translation rules to fix these issues.

Target {target_dialect} Schema (Reference for Exact Table/Column Names and Types):
{schema}

Error Logs:
{error_logs}

Instructions:

Identify common patterns of failure (e.g., date formatting, integer division, quoting rules, NULL handling, ordering differences).
Formulate concise rules to address these failures.
The rules should be instructions for an AI translator (e.g., "When translating X, always do Y").
If a failure seems to be due to ambiguous logic rather than SQL equivalence (e.g., unpredictable sort order), note it but prioritize strict syntax/semantics rules.
Refer to the specific Schema info provided in each error log to understand column types (e.g., if a column is VARCHAR or INT).

Output Format:
Return ONLY the list of rules as a bulleted list. Do not include introductory text.
\end{Verbatim}
\end{tcolorbox}

\section{Error Analysis}
\label{app:error_analysis}

\begin{table*}[t]
\centering
\caption{Post-hoc error distribution of Claude-4.5-Sonnet failures on \textsc{UniQL}. Shares are computed within failed predictions for each dialect and grouped into coarse error categories based on saved SQL predictions, execution errors, execution results, and manual calibration.}
\label{tab:bird_style_error_taxonomy}
\small
\setlength{\tabcolsep}{3.6pt}
\begin{tabular}{lrrrrrr}
\toprule
Dialect & Fail & Syntax/func. & Schema/ref. & Value/filter & SQL logic & Other \\
\midrule
ClickHouse & 662 & 4.53 & 27.19 & 46.98 & 21.15 & 0.15 \\
Doris & 725 & 12.83 & 24.14 & 42.62 & 20.28 & 0.14 \\
Drill & 732 & 13.80 & 22.13 & 16.26 & 47.81 & 0.00 \\
Druid & 922 & 56.40 & 12.15 & 22.45 & 8.68 & 0.33 \\
DuckDB & 686 & 7.87 & 30.32 & 41.69 & 19.97 & 0.15 \\
Hive & 620 & 0.81 & 29.03 & 47.10 & 23.06 & 0.00 \\
MySQL & 636 & 4.87 & 28.77 & 45.13 & 20.91 & 0.31 \\
Oracle & 556 & 9.89 & 43.88 & 33.45 & 12.05 & 0.72 \\
PostgreSQL & 691 & 2.03 & 37.77 & 38.93 & 21.13 & 0.14 \\
Presto & 643 & 2.64 & 28.15 & 48.99 & 19.91 & 0.31 \\
Spark & 673 & 1.78 & 29.27 & 46.81 & 22.14 & 0.00 \\
SQLite & 616 & 0.65 & 28.57 & 45.62 & 24.68 & 0.49 \\
StarRocks & 674 & 6.97 & 25.82 & 45.70 & 21.51 & 0.00 \\
T-SQL & 671 & 8.49 & 25.93 & 44.71 & 20.72 & 0.15 \\
Teradata & 955 & 57.38 & 10.37 & 21.36 & 10.89 & 0.00 \\
Trino & 674 & 2.52 & 30.71 & 47.92 & 18.84 & 0.00 \\
\bottomrule
\end{tabular}
\end{table*}

To further understand the remaining failures of strong models on \textsc{UniQL}, we conduct a post-hoc error analysis for Claude-4.5-Sonnet, the best-performing model in our main experiments. We assign each failed prediction to one of five coarse categories: \emph{syntax/function}, \emph{schema/reference}, \emph{value/filter}, \emph{SQL logic}, and \emph{other}. The analysis is based on saved model predictions, execution errors, execution results, and manual calibration.

\paragraph{Error categories.}
\emph{Syntax/function} errors refer to failures caused by invalid dialect-specific syntax, unsupported functions or operators, quoting and casting problems, or database-engine-specific execution errors. 
\emph{Schema/reference} errors include wrong table or column references, incorrect joins, aliasing mistakes, or selecting attributes from an inappropriate relation. 
\emph{Value/filter} errors capture incorrect literals, missing or spurious predicates, NULL-sensitive conditions, boundary cases, cardinality-changing filters, and tie-breaking differences. 
\emph{SQL logic} errors denote executable or nearly executable queries whose structure is semantically incomplete or incorrect, including missing aggregation, ranking, grouping, nesting, ordering scope, or required answer columns. 
The \emph{other} category covers rare failures that do not fit the above groups.

\paragraph{Dialect-level error distribution.}
Table~\ref{tab:bird_style_error_taxonomy} shows that Claude-4.5-Sonnet's failure modes are highly dialect-dependent. For most dialects, value/filter and schema/reference errors dominate, suggesting that the model often produces plausible SQL skeletons but fails to precisely align predicates, values, joins, or output attributes with the intended query. For example, Presto, Trino, Hive, ClickHouse, Spark, SQLite, and MySQL all have value/filter errors as the largest category. This indicates that many failures are not purely syntactic; instead, they arise from subtle mismatches in literal values, filters, NULL handling, boundary conditions, tie-breaking, or result cardinality.

In contrast, Druid and Teradata exhibit much larger syntax/function error shares, reaching 56.40\% and 57.38\% of failed predictions, respectively. This suggests that failures on these systems are more strongly tied to dialect-specific functions, type handling, planner constraints, and execution behavior. Drill shows another distinct pattern, with SQL logic errors forming the largest portion of failures. This indicates that its failures more often arise from semantically incomplete or structurally incorrect query formulations rather than surface-level syntax alone. These results reinforce the motivation of \textsc{UniQL}: cross-dialect text-to-SQL robustness cannot be inferred from SQLite-only evaluation or a single aggregate accuracy score, because different database systems expose qualitatively different failure modes.

\paragraph{Representative examples.}
Table~\ref{tab:error_examples} provides representative examples for the four major error categories. These examples illustrate how the same broad failure taxonomy captures different kinds of model behavior: dialect-specific execution failures, incorrect schema or relation grounding, subtle value and filter mismatches, and structurally incomplete SQL generation. Long SQL strings are shortened for readability while preserving the key failure pattern.

\begin{table*}[t]
\centering
\caption{Representative examples for the post-hoc error taxonomy. SQL strings are shortened for readability while preserving the key failure pattern.}
\label{tab:error_examples}
\small
\setlength{\tabcolsep}{3.2pt}
\renewcommand{\arraystretch}{1.12}
\begin{tabular}{p{0.14\textwidth}p{0.25\textwidth}p{0.26\textwidth}p{0.26\textwidth}}
\toprule
\textbf{Category} & \textbf{Question / Evidence} & \textbf{Predicted SQL} & \textbf{Gold SQL / Key issue} \\
\midrule

\textbf{Syntax/function}
\newline Druid, id=1
&
Please list the lowest three eligible free rates for students aged 5--17 in continuation schools.
\newline
Evidence: eligible free rate = \texttt{Free Meal Count (Ages 5-17)} / \texttt{Enrollment (Ages 5-17)}.
&
\texttt{SELECT "free meal count (ages 5-17)" / "enrollment (ages 5-17)" AS eligible\_free\_rate FROM bird\_california\_schools\_frpm WHERE "educational option type" = 'Continuation' AND "enrollment (ages 5-17)" > 0 ORDER BY eligible\_free\_rate ASC LIMIT 3}
&
The query triggers a Druid planner error:
\texttt{HTTP 400, INVALID\_INPUT. Query could not be planned\ldots}
The gold query uses explicit casting and a \texttt{CASE} expression to handle division and NULL-sensitive values. The prediction also contains a secondary literal mismatch, but the main label is assigned based on the dialect-specific execution failure.
\\
\midrule

\textbf{Schema/reference}
\newline PostgreSQL, id=25
&
Name schools in Riverside for which the average of average math score for SAT is greater than 400, and show the funding type of these schools.
\newline
Evidence: average of average math = sum(average math scores) / count(schools).
&
\texttt{SELECT s.sname, sch.fundingtype FROM satscores s JOIN schools sch ON s.cds = sch.cdscode WHERE s.cname = 'Riverside' AND s.avgscrmath > 400}
&
\texttt{SELECT T1.sname, T2."charter funding type" FROM satscores AS T1 INNER JOIN frpm AS T2 ON T1.cds = T2.cdscode WHERE T2."district name" LIKE 'Riverside\%' GROUP BY T1.sname, T2."charter funding type" HAVING CAST(SUM(T1.avgscrmath) AS real) / COUNT(T1.cds) > 400}
\newline
The prediction selects the funding type from the wrong relation and misses the required reference to \texttt{frpm}. It also has a secondary aggregation-condition mismatch.
\\
\midrule

\textbf{Value/filter}
\newline Oracle, id=30
&
Which cities have the top 5 lowest enrollment number for students in grades 1 through 12?
\newline
Evidence: K--12 refers to students in grades 1 through 12.
&
\texttt{SELECT s.CITY FROM SCHOOLS s JOIN FRPM f ON s.CDSCODE = f.CDSCODE WHERE f."ENROLLMENT (K-12)" IS NOT NULL AND s.CITY IS NOT NULL GROUP BY s.CITY ORDER BY SUM(f."ENROLLMENT (K-12)") ASC FETCH FIRST 5 ROWS ONLY}
&
\texttt{SELECT T2.CITY FROM FRPM T1 INNER JOIN SCHOOLS T2 ON T1.CDSCODE = T2.CDSCODE GROUP BY T2.CITY ORDER BY SUM(NVL(T1."ENROLLMENT (K-12)",0)) ASC, T2.CITY ASC FETCH FIRST 5 ROWS ONLY}
\newline
The prediction follows the correct high-level aggregation pattern, but changes NULL-sensitive filtering by removing rows with NULL enrollment or city values. The gold query instead treats NULL enrollment as 0 and uses city-level tie-breaking, which can change the top-5 returned cities.
\\
\midrule

\textbf{SQL logic}
\newline SQLite, id=17
&
Rank schools by their average score in Writing where the score is greater than 499, showing their charter numbers.
\newline
Evidence: valid charter number means the number is not null.
&
\texttt{SELECT T2.CharterNum FROM satscores AS T1 INNER JOIN schools AS T2 ON T1.cds = T2.CDSCode WHERE T2.CharterNum IS NOT NULL AND T1.AvgScrWrite > 499 ORDER BY T1.AvgScrWrite DESC}
&
\texttt{SELECT CharterNum, AvgScrWrite, RANK() OVER (ORDER BY AvgScrWrite DESC) AS WritingScoreRank FROM schools AS T1 INNER JOIN satscores AS T2 ON T1.CDSCode = T2.cds WHERE T2.AvgScrWrite > 499 AND CharterNum IS NOT NULL}
\newline
The prediction returns only the charter number and misses both the writing score and the ranking computation, leading to an incomplete answer shape.
\\
\bottomrule
\end{tabular}
\end{table*}



\section{Detailed Stratified Results by Dialect}
\label{app:detailed_stratified_results}

To complement the aggregate stratified analysis in Section~\ref{sec:stratified_analysis}, we report the full dialect-level results under the two partitions used in the main text. The first partition follows the original BIRD difficulty labels, separating questions into simple, moderate, and challenging subsets. The second follows the \textsc{UniQL} construction source, grouping target SQL annotations by whether they are obtained from the SQLGlot tool stage, LLM translation, self-reflection, rule refinement, or human verification. Table~\ref{tab:appendix_construction_statistics} summarizes the subset sizes for these two stratification schemes, while the following tables provide execution accuracy for every evaluated model on each SQL dialect. These detailed results expose whether an aggregate performance trend is stable across dialects or driven by particular database systems and construction paths.

\begin{table*}[!t]
\centering
\scriptsize
\setlength{\tabcolsep}{3.2pt}
\caption{Construction statistics for the dialects in \textsc{UniQL}  under the two stratification schemes used in the analysis. For each dialect, both the construction-source block and the BIRD-difficulty block partition the same aligned split of 1,534 examples.}
\begin{tabular}{lrrrrrrrr}
\toprule
& \multicolumn{5}{c}{\textsc{UniQL} construction source} & \multicolumn{3}{c}{BIRD difficulty} \\
\cmidrule(lr){2-6}\cmidrule(lr){7-9}
Dialect & Tool & LLM & Reflection & Rule Refinement & Human & Simple & Moderate & Challenging \\
\midrule
ClickHouse & 1300 & 142 & 63 & 12 & 17 & 925 & 464 & 145 \\
Doris & 1424 & 39 & 48 & 10 & 13 & 925 & 464 & 145 \\
Drill & 1344 & 0 & 2 & 148 & 40 & 925 & 464 & 145 \\
Druid & 1145 & 304 & 17 & 44 & 24 & 925 & 464 & 145 \\
DuckDB & 1369 & 52 & 31 & 10 & 72 & 925 & 464 & 145 \\
Hive & 959 & 195 & 151 & 100 & 129 & 925 & 464 & 145 \\
MySQL & 1472 & 20 & 26 & 6 & 10 & 925 & 464 & 145 \\
Oracle & 1166 & 157 & 164 & 22 & 25 & 925 & 464 & 145 \\
PostgreSQL & 1282 & 151 & 57 & 24 & 20 & 925 & 464 & 145 \\
Presto & 1299 & 184 & 17 & 14 & 20 & 925 & 464 & 145 \\
Spark & 1344 & 68 & 76 & 10 & 36 & 925 & 464 & 145 \\
SQLite & -- & -- & -- & -- & -- & 925 & 464 & 145 \\
StarRocks & 1319 & 122 & 63 & 12 & 18 & 925 & 464 & 145 \\
T-SQL & 1368 & 106 & 35 & 11 & 14 & 925 & 464 & 145 \\
Teradata & 828 & 388 & 218 & 39 & 61 & 925 & 464 & 145 \\
Trino & 1377 & 83 & 51 & 6 & 17 & 925 & 464 & 145 \\
\bottomrule
\end{tabular}
\label{tab:appendix_construction_statistics}
\end{table*}

\begin{table*}[t]
\centering
\caption{Execution accuracy (\%) on ClickHouse under the original BIRD difficulty and \textsc{UniQL} construction-source stratifications.}
\label{tab:appendix_difficulty_clickhouse}
\scriptsize
\setlength{\tabcolsep}{2.2pt}
\begin{tabular}{lrrrrrrrrr}
\toprule
& & \multicolumn{3}{c}{BIRD difficulty} & \multicolumn{5}{c}{\textsc{UniQL} construction source} \\
\cmidrule(lr){3-5}\cmidrule(lr){6-10}
Model & All & Simple & Moderate & Challenging & Tool & LLM & Reflection & Rule Refinement & Human \\
\midrule
\multicolumn{10}{l}{\textit{Closed-weight LLMs}} \\
GPT-3.5-Turbo & 35.14 & 43.78 & 22.63 & 20.00 & 37.69 & 28.87 & 11.11 & 8.33 & 0.00 \\
GPT-5-mini & 51.56 & 58.38 & 41.81 & 39.31 & 53.77 & 52.11 & 23.81 & 8.33 & 11.76 \\
GPT-5.1-codex & 52.61 & 59.14 & 43.75 & 39.31 & 55.54 & 51.41 & 12.70 & 8.33 & 17.65 \\
Gemini-2.5-Pro & 53.98 & 61.30 & 43.75 & 40.00 & 57.00 & 52.11 & 17.46 & 8.33 & 5.88 \\
Claude-4.5-Sonnet & 56.84 & 63.78 & 48.06 & 40.69 & 59.85 & 55.63 & 19.05 & 8.33 & 11.76 \\
\midrule
\multicolumn{10}{l}{\textit{Open-weight LLMs}} \\
Qwen3-1.7B & 33.90 & 42.16 & 23.06 & 15.86 & 36.38 & 28.87 & 7.94 & 0.00 & 5.88 \\
Qwen3-4B & 43.02 & 51.78 & 31.90 & 22.76 & 46.62 & 35.21 & 4.76 & 0.00 & 5.88 \\
Qwen3-8B & 46.54 & 55.24 & 36.42 & 23.45 & 50.23 & 37.32 & 7.94 & 0.00 & 17.65 \\
Qwen3-32B & 49.74 & 56.54 & 41.38 & 33.10 & 52.77 & 45.07 & 15.87 & 0.00 & 17.65 \\
Llama-3-8B-Inst & 20.80 & 25.84 & 14.22 & 9.66 & 22.92 & 13.38 & 3.17 & 0.00 & 0.00 \\
Llama-3-70B-Inst & 40.16 & 48.76 & 30.39 & 16.55 & 43.31 & 32.39 & 9.52 & 0.00 & 5.88 \\
DeepSeek-Coder-16B & 32.46 & 39.57 & 21.77 & 21.38 & 35.54 & 19.72 & 9.52 & 8.33 & 5.88 \\
DeepSeek-v4-flash & 48.04 & 55.78 & 38.58 & 28.97 & 51.69 & 40.14 & 11.11 & 0.00 & 5.88 \\
\bottomrule
\end{tabular}
\end{table*}

\begin{table*}[t]
\centering
\caption{Execution accuracy (\%) on Doris under the original BIRD difficulty and \textsc{UniQL} construction-source stratifications.}
\label{tab:appendix_difficulty_doris}
\scriptsize
\setlength{\tabcolsep}{2.2pt}
\begin{tabular}{lrrrrrrrrr}
\toprule
& & \multicolumn{3}{c}{BIRD difficulty} & \multicolumn{5}{c}{\textsc{UniQL} construction source} \\
\cmidrule(lr){3-5}\cmidrule(lr){6-10}
Model & All & Simple & Moderate & Challenging & Tool & LLM & Reflection & Rule Refinement & Human \\
\midrule
\multicolumn{10}{l}{\textit{Closed-weight LLMs}} \\
GPT-3.5-Turbo & 38.33 & 46.59 & 25.65 & 26.21 & 40.52 & 23.08 & 2.08 & 0.00 & 7.69 \\
GPT-5-mini & 49.80 & 56.65 & 40.52 & 35.86 & 51.62 & 41.03 & 18.75 & 0.00 & 30.77 \\
GPT-5.1-codex & 49.93 & 56.11 & 41.59 & 37.24 & 51.69 & 48.72 & 16.67 & 0.00 & 23.08 \\
Gemini-2.5-Pro & 51.76 & 57.84 & 44.40 & 36.55 & 53.93 & 46.15 & 10.42 & 0.00 & 23.08 \\
Claude-4.5-Sonnet & 52.74 & 58.59 & 45.69 & 37.93 & 54.42 & 58.97 & 14.58 & 10.00 & 23.08 \\
\midrule
\multicolumn{10}{l}{\textit{Open-weight LLMs}} \\
Qwen3-1.7B & 35.07 & 43.89 & 23.71 & 15.17 & 36.73 & 30.77 & 2.08 & 0.00 & 15.38 \\
Qwen3-4B & 44.52 & 51.68 & 35.56 & 27.59 & 46.49 & 46.15 & 2.08 & 0.00 & 15.38 \\
Qwen3-8B & 46.15 & 52.86 & 38.36 & 28.28 & 48.10 & 41.03 & 10.42 & 0.00 & 15.38 \\
Qwen3-32B & 47.20 & 53.51 & 39.66 & 31.03 & 48.88 & 56.41 & 8.33 & 0.00 & 15.38 \\
Llama-3-8B-Inst & 23.99 & 30.05 & 17.03 & 7.59 & 25.42 & 12.82 & 2.08 & 0.00 & 0.00 \\
Llama-3-70B-Inst & 40.09 & 47.89 & 30.60 & 20.69 & 41.92 & 33.33 & 6.25 & 0.00 & 15.38 \\
DeepSeek-Coder-16B & 31.10 & 38.92 & 20.69 & 14.48 & 32.87 & 17.95 & 2.08 & 0.00 & 7.69 \\
DeepSeek-v4-flash & 46.94 & 53.08 & 40.09 & 29.66 & 48.67 & 51.28 & 8.33 & 0.00 & 23.08 \\
\bottomrule
\end{tabular}
\end{table*}

\begin{table*}[t]
\centering
\caption{Execution accuracy (\%) on Drill under the original BIRD difficulty and \textsc{UniQL} construction-source stratifications.}
\label{tab:appendix_difficulty_drill}
\scriptsize
\setlength{\tabcolsep}{2.2pt}
\begin{tabular}{lrrrrrrrrr}
\toprule
& & \multicolumn{3}{c}{BIRD difficulty} & \multicolumn{5}{c}{\textsc{UniQL} construction source} \\
\cmidrule(lr){3-5}\cmidrule(lr){6-10}
Model & All & Simple & Moderate & Challenging & Tool & LLM & Reflection & Rule Refinement & Human \\
\midrule
\multicolumn{10}{l}{\textit{Closed-weight LLMs}} \\
GPT-3.5-Turbo & 29.20 & 36.86 & 17.89 & 16.55 & 32.22 & -- & 0.00 & 8.78 & 5.00 \\
GPT-5-mini & 48.83 & 56.43 & 39.01 & 31.72 & 51.26 & -- & 50.00 & 36.49 & 12.50 \\
GPT-5.1-codex & 48.11 & 54.38 & 40.95 & 31.03 & 51.19 & -- & 50.00 & 31.76 & 5.00 \\
Gemini-2.5-Pro & 50.85 & 57.51 & 43.10 & 33.10 & 53.94 & -- & 0.00 & 33.78 & 12.50 \\
Claude-4.5-Sonnet & 52.28 & 58.59 & 45.91 & 32.41 & 55.58 & -- & 100.00 & 33.11 & 10.00 \\
\midrule
\multicolumn{10}{l}{\textit{Open-weight LLMs}} \\
Qwen3-1.7B & 27.57 & 35.03 & 17.24 & 13.10 & 29.84 & -- & 0.00 & 13.51 & 5.00 \\
Qwen3-4B & 41.72 & 49.84 & 30.82 & 24.83 & 44.79 & -- & 0.00 & 22.30 & 12.50 \\
Qwen3-8B & 44.13 & 51.78 & 34.27 & 26.90 & 46.88 & -- & 0.00 & 28.38 & 12.50 \\
Qwen3-32B & 46.54 & 53.30 & 39.01 & 27.59 & 49.55 & -- & 50.00 & 29.05 & 10.00 \\
Llama-3-8B-Inst & 18.45 & 24.00 & 11.21 & 6.21 & 20.16 & -- & 50.00 & 6.76 & 2.50 \\
Llama-3-70B-Inst & 37.74 & 45.62 & 28.23 & 17.93 & 40.85 & -- & 0.00 & 18.92 & 5.00 \\
DeepSeek-Coder-16B & 32.07 & 39.68 & 22.41 & 14.48 & 34.38 & -- & 50.00 & 17.57 & 7.50 \\
DeepSeek-v4-flash & 46.28 & 53.84 & 36.21 & 30.34 & 49.26 & -- & 0.00 & 29.05 & 12.50 \\
\bottomrule
\end{tabular}
\end{table*}

\begin{table*}[t]
\centering
\caption{Execution accuracy (\%) on Druid under the original BIRD difficulty and \textsc{UniQL} construction-source stratifications.}
\label{tab:appendix_difficulty_druid}
\scriptsize
\setlength{\tabcolsep}{2.2pt}
\begin{tabular}{lrrrrrrrrr}
\toprule
& & \multicolumn{3}{c}{BIRD difficulty} & \multicolumn{5}{c}{\textsc{UniQL} construction source} \\
\cmidrule(lr){3-5}\cmidrule(lr){6-10}
Model & All & Simple & Moderate & Challenging & Tool & LLM & Reflection & Rule Refinement & Human \\
\midrule
\multicolumn{10}{l}{\textit{Closed-weight LLMs}} \\
GPT-3.5-Turbo & 22.75 & 30.81 & 12.28 & 4.83 & 28.03 & 7.24 & 5.88 & 9.09 & 4.17 \\
GPT-5-mini & 37.74 & 45.30 & 27.80 & 21.38 & 43.14 & 21.71 & 17.65 & 34.09 & 4.17 \\
GPT-5.1-codex & 38.20 & 45.30 & 30.39 & 17.93 & 42.97 & 24.34 & 17.65 & 27.27 & 20.83 \\
Gemini-2.5-Pro & 37.42 & 43.78 & 29.96 & 20.69 & 42.18 & 22.37 & 17.65 & 31.82 & 25.00 \\
Claude-4.5-Sonnet & 39.90 & 47.14 & 30.60 & 23.45 & 45.07 & 24.34 & 11.76 & 31.82 & 25.00 \\
\midrule
\multicolumn{10}{l}{\textit{Open-weight LLMs}} \\
Qwen3-1.7B & 18.71 & 25.95 & 8.62 & 4.83 & 22.97 & 6.58 & 0.00 & 4.55 & 8.33 \\
Qwen3-4B & 26.53 & 34.05 & 17.46 & 7.59 & 31.62 & 9.87 & 5.88 & 22.73 & 16.67 \\
Qwen3-8B & 31.29 & 39.24 & 21.12 & 13.10 & 36.59 & 13.82 & 11.76 & 27.27 & 20.83 \\
Qwen3-32B & 33.83 & 41.08 & 24.78 & 16.55 & 39.91 & 14.14 & 17.65 & 31.82 & 8.33 \\
Llama-3-8B-Inst & 16.56 & 23.35 & 7.11 & 3.45 & 20.17 & 5.92 & 0.00 & 6.82 & 8.33 \\
Llama-3-70B-Inst & 26.92 & 34.92 & 17.89 & 4.83 & 33.10 & 7.89 & 5.88 & 11.36 & 16.67 \\
DeepSeek-Coder-16B & 21.38 & 28.97 & 10.99 & 6.21 & 26.03 & 7.24 & 5.88 & 11.36 & 8.33 \\
DeepSeek-v4-flash & 34.22 & 42.05 & 25.65 & 11.72 & 40.52 & 15.13 & 0.00 & 22.73 & 20.83 \\
\bottomrule
\end{tabular}
\end{table*}

\begin{table*}[t]
\centering
\caption{Execution accuracy (\%) on DuckDB under the original BIRD difficulty and \textsc{UniQL} construction-source stratifications.}
\label{tab:appendix_difficulty_duckdb}
\scriptsize
\setlength{\tabcolsep}{2.2pt}
\begin{tabular}{lrrrrrrrrr}
\toprule
& & \multicolumn{3}{c}{BIRD difficulty} & \multicolumn{5}{c}{\textsc{UniQL} construction source} \\
\cmidrule(lr){3-5}\cmidrule(lr){6-10}
Model & All & Simple & Moderate & Challenging & Tool & LLM & Reflection & Rule Refinement & Human \\
\midrule
\multicolumn{10}{l}{\textit{Closed-weight LLMs}} \\
GPT-3.5-Turbo & 37.09 & 46.27 & 22.41 & 25.52 & 39.30 & 15.38 & 6.45 & 10.00 & 27.78 \\
GPT-5-mini & 52.74 & 58.16 & 44.83 & 43.45 & 54.13 & 40.38 & 29.03 & 10.00 & 51.39 \\
GPT-5.1-codex & 52.93 & 59.57 & 44.40 & 37.93 & 54.86 & 38.46 & 29.03 & 0.00 & 44.44 \\
Gemini-2.5-Pro & 56.19 & 61.73 & 48.71 & 44.83 & 58.80 & 36.54 & 16.13 & 10.00 & 44.44 \\
Claude-4.5-Sonnet & 55.28 & 61.62 & 47.84 & 38.62 & 57.05 & 46.15 & 25.81 & 10.00 & 47.22 \\
\midrule
\multicolumn{10}{l}{\textit{Open-weight LLMs}} \\
Qwen3-1.7B & 35.27 & 42.81 & 25.43 & 18.62 & 37.91 & 13.46 & 3.23 & 10.00 & 18.06 \\
Qwen3-4B & 44.26 & 51.78 & 35.13 & 25.52 & 47.33 & 23.08 & 6.45 & 10.00 & 22.22 \\
Qwen3-8B & 47.00 & 55.14 & 37.72 & 24.83 & 49.96 & 23.08 & 12.90 & 10.00 & 27.78 \\
Qwen3-32B & 49.93 & 56.32 & 42.24 & 33.79 & 52.37 & 40.38 & 12.90 & 10.00 & 31.94 \\
Llama-3-8B-Inst & 22.29 & 27.57 & 15.95 & 8.97 & 24.11 & 5.77 & 0.00 & 0.00 & 12.50 \\
Llama-3-70B-Inst & 41.72 & 49.84 & 32.97 & 17.93 & 44.56 & 21.15 & 3.23 & 10.00 & 23.61 \\
DeepSeek-Coder-16B & 32.72 & 40.54 & 22.41 & 15.86 & 34.70 & 11.54 & 6.45 & 10.00 & 25.00 \\
DeepSeek-v4-flash & 49.02 & 55.68 & 41.38 & 31.03 & 51.35 & 34.62 & 9.68 & 10.00 & 37.50 \\
\bottomrule
\end{tabular}
\end{table*}

\begin{table*}[t]
\centering
\caption{Execution accuracy (\%) on Hive under the original BIRD difficulty and \textsc{UniQL} construction-source stratifications.}
\label{tab:appendix_difficulty_hive}
\scriptsize
\setlength{\tabcolsep}{2.2pt}
\begin{tabular}{lrrrrrrrrr}
\toprule
& & \multicolumn{3}{c}{BIRD difficulty} & \multicolumn{5}{c}{\textsc{UniQL} construction source} \\
\cmidrule(lr){3-5}\cmidrule(lr){6-10}
Model & All & Simple & Moderate & Challenging & Tool & LLM & Reflection & Rule Refinement & Human \\
\midrule
\multicolumn{10}{l}{\textit{Closed-weight LLMs}} \\
GPT-3.5-Turbo & 36.18 & 45.41 & 22.41 & 21.38 & 44.94 & 30.26 & 10.60 & 3.00 & 35.66 \\
GPT-5-mini & 54.04 & 59.89 & 45.04 & 45.52 & 57.87 & 48.72 & 50.33 & 30.00 & 56.59 \\
GPT-5.1-codex & 55.48 & 60.65 & 47.84 & 46.90 & 59.02 & 50.26 & 51.66 & 35.00 & 57.36 \\
Gemini-2.5-Pro & 59.32 & 65.30 & 51.08 & 47.59 & 61.84 & 55.90 & 54.30 & 47.00 & 61.24 \\
Claude-4.5-Sonnet & 59.58 & 65.51 & 51.72 & 46.90 & 63.82 & 54.87 & 54.30 & 37.00 & 58.91 \\
\midrule
\multicolumn{10}{l}{\textit{Open-weight LLMs}} \\
Qwen3-1.7B & 36.70 & 45.19 & 24.57 & 21.38 & 43.07 & 24.62 & 23.18 & 19.00 & 37.21 \\
Qwen3-4B & 46.74 & 55.46 & 34.70 & 29.66 & 52.24 & 38.46 & 35.10 & 25.00 & 48.84 \\
Qwen3-8B & 44.85 & 52.86 & 35.34 & 24.14 & 53.39 & 38.97 & 23.84 & 7.00 & 44.19 \\
Qwen3-32B & 51.96 & 59.03 & 42.89 & 35.86 & 56.10 & 46.67 & 43.05 & 32.00 & 55.04 \\
Llama-3-8B-Inst & 22.23 & 28.97 & 13.36 & 7.59 & 25.65 & 12.31 & 15.23 & 17.00 & 24.03 \\
Llama-3-70B-Inst & 42.63 & 51.46 & 31.47 & 22.07 & 47.65 & 35.38 & 28.48 & 25.00 & 46.51 \\
DeepSeek-Coder-16B & 34.88 & 42.92 & 22.84 & 22.07 & 39.52 & 24.62 & 27.81 & 17.00 & 37.98 \\
DeepSeek-v4-flash & 52.80 & 59.03 & 44.40 & 40.00 & 57.46 & 43.08 & 42.38 & 35.00 & 58.91 \\
\bottomrule
\end{tabular}
\end{table*}

\begin{table*}[t]
\centering
\caption{Execution accuracy (\%) on MySQL under the original BIRD difficulty and \textsc{UniQL} construction-source stratifications.}
\label{tab:appendix_difficulty_mysql}
\scriptsize
\setlength{\tabcolsep}{2.2pt}
\begin{tabular}{lrrrrrrrrr}
\toprule
& & \multicolumn{3}{c}{BIRD difficulty} & \multicolumn{5}{c}{\textsc{UniQL} construction source} \\
\cmidrule(lr){3-5}\cmidrule(lr){6-10}
Model & All & Simple & Moderate & Challenging & Tool & LLM & Reflection & Rule Refinement & Human \\
\midrule
\multicolumn{10}{l}{\textit{Closed-weight LLMs}} \\
GPT-3.5-Turbo & 40.81 & 50.49 & 27.59 & 21.38 & 42.05 & 20.00 & 7.69 & 0.00 & 10.00 \\
GPT-5-mini & 54.30 & 60.43 & 45.47 & 43.45 & 55.37 & 25.00 & 30.77 & 0.00 & 50.00 \\
GPT-5.1-codex & 54.89 & 61.95 & 45.26 & 40.69 & 56.05 & 30.00 & 26.92 & 0.00 & 40.00 \\
Gemini-2.5-Pro & 57.82 & 64.00 & 50.22 & 42.76 & 59.38 & 30.00 & 19.23 & 0.00 & 20.00 \\
Claude-4.5-Sonnet & 58.54 & 65.73 & 49.78 & 40.69 & 59.85 & 35.00 & 23.08 & 0.00 & 40.00 \\
\midrule
\multicolumn{10}{l}{\textit{Open-weight LLMs}} \\
Qwen3-1.7B & 35.85 & 45.73 & 23.28 & 13.10 & 37.09 & 5.00 & 3.85 & 0.00 & 20.00 \\
Qwen3-4B & 46.61 & 56.00 & 33.84 & 27.59 & 47.89 & 15.00 & 15.38 & 0.00 & 30.00 \\
Qwen3-8B & 50.20 & 58.27 & 40.52 & 29.66 & 51.43 & 20.00 & 23.08 & 0.00 & 30.00 \\
Qwen3-32B & 52.02 & 59.46 & 42.67 & 34.48 & 53.46 & 25.00 & 15.38 & 0.00 & 20.00 \\
Llama-3-8B-Inst & 22.03 & 28.86 & 13.15 & 6.90 & 22.83 & 0.00 & 3.85 & 0.00 & 10.00 \\
Llama-3-70B-Inst & 43.74 & 53.08 & 32.54 & 20.00 & 45.11 & 20.00 & 3.85 & 0.00 & 20.00 \\
DeepSeek-Coder-16B & 36.57 & 45.30 & 26.29 & 13.79 & 37.57 & 15.00 & 11.54 & 0.00 & 20.00 \\
DeepSeek-v4-flash & 52.54 & 60.22 & 43.32 & 33.10 & 53.87 & 20.00 & 30.77 & 0.00 & 10.00 \\
\bottomrule
\end{tabular}
\end{table*}

\begin{table*}[t]
\centering
\caption{Execution accuracy (\%) on Oracle under the original BIRD difficulty and \textsc{UniQL} construction-source stratifications.}
\label{tab:appendix_difficulty_oracle}
\scriptsize
\setlength{\tabcolsep}{2.2pt}
\begin{tabular}{lrrrrrrrrr}
\toprule
& & \multicolumn{3}{c}{BIRD difficulty} & \multicolumn{5}{c}{\textsc{UniQL} construction source} \\
\cmidrule(lr){3-5}\cmidrule(lr){6-10}
Model & All & Simple & Moderate & Challenging & Tool & LLM & Reflection & Rule Refinement & Human \\
\midrule
\multicolumn{10}{l}{\textit{Closed-weight LLMs}} \\
GPT-3.5-Turbo & 49.35 & 59.57 & 37.50 & 22.07 & 60.12 & 17.83 & 12.20 & 13.64 & 20.00 \\
GPT-5-mini & 60.56 & 64.86 & 54.09 & 53.79 & 69.98 & 42.68 & 21.95 & 18.18 & 24.00 \\
GPT-5.1-codex & 49.67 & 54.81 & 42.46 & 40.00 & 57.80 & 33.12 & 15.85 & 18.18 & 24.00 \\
Gemini-2.5-Pro & 34.29 & 39.46 & 25.65 & 28.97 & 39.71 & 24.84 & 11.59 & 9.09 & 12.00 \\
Claude-4.5-Sonnet & 63.75 & 69.51 & 56.47 & 50.34 & 74.70 & 42.04 & 17.68 & 22.73 & 28.00 \\
\midrule
\multicolumn{10}{l}{\textit{Open-weight LLMs}} \\
Qwen3-1.7B & 43.48 & 52.65 & 31.47 & 23.45 & 53.86 & 12.74 & 10.37 & 4.55 & 4.00 \\
Qwen3-4B & 44.92 & 50.27 & 38.79 & 30.34 & 54.29 & 19.11 & 12.20 & 13.64 & 12.00 \\
Qwen3-8B & 51.11 & 56.32 & 45.69 & 35.17 & 60.38 & 29.30 & 14.63 & 18.18 & 24.00 \\
Qwen3-32B & 53.06 & 58.70 & 46.55 & 37.93 & 62.35 & 33.76 & 16.46 & 9.09 & 20.00 \\
Llama-3-8B-Inst & 31.23 & 39.57 & 20.04 & 13.79 & 38.68 & 7.01 & 8.54 & 4.55 & 8.00 \\
Llama-3-70B-Inst & 51.56 & 59.24 & 43.97 & 26.90 & 63.12 & 18.47 & 11.59 & 4.55 & 24.00 \\
DeepSeek-Coder-16B & 45.44 & 54.81 & 33.84 & 22.76 & 55.75 & 14.65 & 11.59 & 9.09 & 12.00 \\
DeepSeek-v4-flash & 57.50 & 64.54 & 49.14 & 39.31 & 68.27 & 32.48 & 15.85 & 18.18 & 20.00 \\
\bottomrule
\end{tabular}
\end{table*}

\begin{table*}[t]
\centering
\caption{Execution accuracy (\%) on PostgreSQL under the original BIRD difficulty and \textsc{UniQL} construction-source stratifications.}
\label{tab:appendix_difficulty_postgresql}
\scriptsize
\setlength{\tabcolsep}{2.2pt}
\begin{tabular}{lrrrrrrrrr}
\toprule
& & \multicolumn{3}{c}{BIRD difficulty} & \multicolumn{5}{c}{\textsc{UniQL} construction source} \\
\cmidrule(lr){3-5}\cmidrule(lr){6-10}
Model & All & Simple & Moderate & Challenging & Tool & LLM & Reflection & Rule Refinement & Human \\
\midrule
\multicolumn{10}{l}{\textit{Closed-weight LLMs}} \\
GPT-3.5-Turbo & 36.44 & 45.73 & 23.49 & 18.62 & 41.50 & 12.58 & 12.28 & 4.17 & 0.00 \\
GPT-5-mini & 51.37 & 58.05 & 40.52 & 43.45 & 54.99 & 38.41 & 29.82 & 12.50 & 25.00 \\
GPT-5.1-codex & 52.35 & 59.46 & 42.46 & 38.62 & 56.40 & 37.09 & 29.82 & 8.33 & 25.00 \\
Gemini-2.5-Pro & 53.59 & 59.24 & 45.69 & 42.76 & 58.19 & 39.07 & 22.81 & 8.33 & 10.00 \\
Claude-4.5-Sonnet & 54.95 & 60.76 & 47.41 & 42.07 & 58.81 & 42.38 & 35.09 & 12.50 & 10.00 \\
\midrule
\multicolumn{10}{l}{\textit{Open-weight LLMs}} \\
Qwen3-1.7B & 32.92 & 41.62 & 20.91 & 15.86 & 36.97 & 14.57 & 12.28 & 4.17 & 5.00 \\
Qwen3-4B & 42.70 & 50.27 & 32.33 & 27.59 & 47.74 & 19.21 & 19.30 & 4.17 & 10.00 \\
Qwen3-8B & 46.35 & 54.49 & 36.85 & 24.83 & 50.39 & 29.80 & 26.32 & 8.33 & 15.00 \\
Qwen3-32B & 48.50 & 56.22 & 39.01 & 29.66 & 52.34 & 34.44 & 29.82 & 4.17 & 15.00 \\
Llama-3-8B-Inst & 22.43 & 28.97 & 15.09 & 4.14 & 25.66 & 6.62 & 8.77 & 0.00 & 0.00 \\
Llama-3-70B-Inst & 40.16 & 48.54 & 31.25 & 15.17 & 45.09 & 17.88 & 15.79 & 4.17 & 5.00 \\
DeepSeek-Coder-16B & 31.88 & 39.35 & 21.34 & 17.93 & 36.12 & 13.91 & 5.26 & 4.17 & 5.00 \\
DeepSeek-v4-flash & 48.50 & 55.78 & 39.87 & 29.66 & 52.65 & 32.45 & 28.07 & 8.33 & 10.00 \\
\bottomrule
\end{tabular}
\end{table*}

\begin{table*}[t]
\centering
\caption{Execution accuracy (\%) on Presto under the original BIRD difficulty and \textsc{UniQL} construction-source stratifications.}
\label{tab:appendix_difficulty_presto}
\scriptsize
\setlength{\tabcolsep}{2.2pt}
\begin{tabular}{lrrrrrrrrr}
\toprule
& & \multicolumn{3}{c}{BIRD difficulty} & \multicolumn{5}{c}{\textsc{UniQL} construction source} \\
\cmidrule(lr){3-5}\cmidrule(lr){6-10}
Model & All & Simple & Moderate & Challenging & Tool & LLM & Reflection & Rule Refinement & Human \\
\midrule
\multicolumn{10}{l}{\textit{Closed-weight LLMs}} \\
GPT-3.5-Turbo & 36.31 & 47.03 & 20.69 & 17.93 & 40.72 & 13.59 & 11.76 & 0.00 & 5.00 \\
GPT-5-mini & 50.98 & 57.30 & 42.67 & 37.24 & 53.50 & 39.13 & 35.29 & 28.57 & 25.00 \\
GPT-5.1-codex & 52.41 & 59.14 & 43.32 & 38.62 & 54.97 & 39.67 & 41.18 & 28.57 & 30.00 \\
Gemini-2.5-Pro & 57.89 & 63.78 & 51.08 & 42.07 & 61.28 & 39.67 & 47.06 & 42.86 & 25.00 \\
Claude-4.5-Sonnet & 58.08 & 64.11 & 50.00 & 45.52 & 60.66 & 45.65 & 52.94 & 28.57 & 30.00 \\
\midrule
\multicolumn{10}{l}{\textit{Open-weight LLMs}} \\
Qwen3-1.7B & 32.46 & 41.08 & 21.55 & 12.41 & 36.87 & 7.61 & 11.76 & 7.14 & 10.00 \\
Qwen3-4B & 41.72 & 51.35 & 28.88 & 21.38 & 46.81 & 13.59 & 17.65 & 0.00 & 20.00 \\
Qwen3-8B & 32.99 & 38.27 & 28.23 & 14.48 & 36.57 & 13.04 & 29.41 & 0.00 & 10.00 \\
Qwen3-32B & 47.78 & 56.65 & 38.15 & 22.07 & 51.04 & 32.07 & 35.29 & 14.29 & 15.00 \\
Llama-3-8B-Inst & 22.23 & 29.41 & 12.50 & 7.59 & 25.02 & 6.52 & 11.76 & 0.00 & 10.00 \\
Llama-3-70B-Inst & 38.01 & 47.35 & 26.72 & 14.48 & 43.11 & 10.33 & 11.76 & 0.00 & 10.00 \\
DeepSeek-Coder-16B & 31.55 & 39.78 & 20.47 & 14.48 & 35.33 & 11.41 & 17.65 & 0.00 & 5.00 \\
DeepSeek-v4-flash & 49.02 & 57.41 & 39.44 & 26.21 & 53.12 & 26.63 & 41.18 & 7.14 & 25.00 \\
\bottomrule
\end{tabular}
\end{table*}

\begin{table*}[t]
\centering
\caption{Execution accuracy (\%) on Spark under the original BIRD difficulty and \textsc{UniQL} construction-source stratifications.}
\label{tab:appendix_difficulty_spark}
\scriptsize
\setlength{\tabcolsep}{2.2pt}
\begin{tabular}{lrrrrrrrrr}
\toprule
& & \multicolumn{3}{c}{BIRD difficulty} & \multicolumn{5}{c}{\textsc{UniQL} construction source} \\
\cmidrule(lr){3-5}\cmidrule(lr){6-10}
Model & All & Simple & Moderate & Challenging & Tool & LLM & Reflection & Rule Refinement & Human \\
\midrule
\multicolumn{10}{l}{\textit{Closed-weight LLMs}} \\
GPT-3.5-Turbo & 39.37 & 48.86 & 26.08 & 21.38 & 43.15 & 22.06 & 3.95 & 0.00 & 16.67 \\
GPT-5-mini & 52.93 & 57.95 & 45.26 & 45.52 & 56.85 & 35.29 & 11.84 & 0.00 & 41.67 \\
GPT-5.1-codex & 53.65 & 59.35 & 45.04 & 44.83 & 57.59 & 38.24 & 13.16 & 0.00 & 36.11 \\
Gemini-2.5-Pro & 56.65 & 61.84 & 49.57 & 46.21 & 60.71 & 47.06 & 13.16 & 0.00 & 30.56 \\
Claude-4.5-Sonnet & 56.13 & 62.16 & 47.84 & 44.14 & 60.27 & 44.12 & 15.79 & 0.00 & 25.00 \\
\midrule
\multicolumn{10}{l}{\textit{Open-weight LLMs}} \\
Qwen3-1.7B & 35.27 & 42.81 & 25.00 & 20.00 & 38.62 & 23.53 & 2.63 & 0.00 & 11.11 \\
Qwen3-4B & 43.94 & 52.43 & 32.11 & 27.59 & 47.40 & 35.29 & 3.95 & 0.00 & 27.78 \\
Qwen3-8B & 47.46 & 54.92 & 38.36 & 28.97 & 50.82 & 19.71 & 10.53 & 0.00 & 27.78 \\
Qwen3-32B & 49.22 & 56.43 & 40.73 & 30.34 & 52.98 & 39.71 & 7.89 & 0.00 & 27.78 \\
Llama-3-8B-Inst & 23.34 & 29.84 & 14.87 & 8.97 & 25.97 & 5.88 & 1.32 & 0.00 & 11.11 \\
Llama-3-70B-Inst & 42.89 & 50.70 & 34.05 & 21.38 & 46.95 & 26.47 & 2.63 & 0.00 & 19.44 \\
DeepSeek-Coder-16B & 34.42 & 42.05 & 24.35 & 17.93 & 37.95 & 17.65 & 3.95 & 0.00 & 8.33 \\
DeepSeek-v4-flash & 51.56 & 57.84 & 43.53 & 37.24 & 55.88 & 36.76 & 9.21 & 0.00 & 22.22 \\
\bottomrule
\end{tabular}
\end{table*}

\begin{table*}[t]
\centering
\caption{Execution accuracy (\%) on SQLite under the original BIRD difficulty and \textsc{UniQL} construction-source stratifications.}
\label{tab:appendix_difficulty_sqlite}
\scriptsize
\setlength{\tabcolsep}{2.2pt}
\begin{tabular}{lrrrrrrrrr}
\toprule
& & \multicolumn{3}{c}{BIRD difficulty} & \multicolumn{5}{c}{\textsc{UniQL} construction source} \\
\cmidrule(lr){3-5}\cmidrule(lr){6-10}
Model & All & Simple & Moderate & Challenging & Tool & LLM & Reflection & Rule Refinement & Human \\
\midrule
\multicolumn{10}{l}{\textit{Closed-weight LLMs}} \\
GPT-3.5-Turbo & 41.13 & 50.81 & 27.16 & 24.14 & 47.45 & 31.28 & 31.13 & 29.00 & 30.23 \\
GPT-5-mini & 52.48 & 57.95 & 44.40 & 43.45 & 54.61 & 42.65 & 38.16 & 20.00 & 30.56 \\
GPT-5.1-codex & 53.78 & 59.46 & 45.91 & 42.76 & 56.47 & 38.24 & 36.84 & 20.00 & 27.78 \\
Gemini-2.5-Pro & 59.78 & 65.41 & 52.16 & 48.28 & 61.53 & 57.35 & 48.68 & 40.00 & 27.78 \\
Claude-4.5-Sonnet & 59.84 & 65.08 & 52.80 & 48.97 & 61.61 & 55.88 & 50.00 & 30.00 & 30.56 \\
\midrule
\multicolumn{10}{l}{\textit{Open-weight LLMs}} \\
Qwen3-1.7B & 35.40 & 43.24 & 24.57 & 20.00 & 37.43 & 20.75 & 17.14 & 9.09 & 14.29 \\
Qwen3-4B & 46.94 & 55.03 & 37.07 & 26.90 & 49.12 & 33.96 & 17.14 & 27.27 & 21.43 \\
Qwen3-8B & 49.09 & 57.41 & 39.01 & 28.28 & 50.73 & 41.51 & 25.71 & 36.36 & 14.29 \\
Qwen3-32B & 53.39 & 60.22 & 45.26 & 35.86 & 55.70 & 42.45 & 17.14 & 27.27 & 21.43 \\
Llama-3-8B-Inst & 23.60 & 29.19 & 17.24 & 8.28 & 24.93 & 13.21 & 20.00 & 0.00 & 0.00 \\
Llama-3-70B-Inst & 42.24 & 50.59 & 32.76 & 19.31 & 44.74 & 23.58 & 17.14 & 9.09 & 28.57 \\
DeepSeek-Coder-16B & 35.01 & 44.32 & 22.63 & 15.17 & 36.77 & 23.58 & 17.14 & 9.09 & 14.29 \\
DeepSeek-v4-flash & 53.46 & 61.51 & 43.97 & 32.41 & 55.65 & 39.71 & 43.42 & 20.00 & 27.78 \\
\bottomrule
\end{tabular}
\end{table*}

\begin{table*}[t]
\centering
\caption{Execution accuracy (\%) on StarRocks under the original BIRD difficulty and \textsc{UniQL} construction-source stratifications.}
\label{tab:appendix_difficulty_starrocks}
\scriptsize
\setlength{\tabcolsep}{2.2pt}
\begin{tabular}{lrrrrrrrrr}
\toprule
& & \multicolumn{3}{c}{BIRD difficulty} & \multicolumn{5}{c}{\textsc{UniQL} construction source} \\
\cmidrule(lr){3-5}\cmidrule(lr){6-10}
Model & All & Simple & Moderate & Challenging & Tool & LLM & Reflection & Rule Refinement & Human \\
\midrule
\multicolumn{10}{l}{\textit{Closed-weight LLMs}} \\
GPT-3.5-Turbo & 39.18 & 47.89 & 25.65 & 26.90 & 43.44 & 18.85 & 4.76 & 0.00 & 11.11 \\
GPT-5-mini & 51.89 & 57.95 & 43.32 & 40.69 & 55.27 & 45.08 & 12.70 & 0.00 & 22.22 \\
GPT-5.1-codex & 52.41 & 57.95 & 45.04 & 40.69 & 55.95 & 41.80 & 17.46 & 0.00 & 22.22 \\
Gemini-2.5-Pro & 54.43 & 61.19 & 44.40 & 43.45 & 57.70 & 49.18 & 12.70 & 0.00 & 33.33 \\
Claude-4.5-Sonnet & 56.06 & 61.95 & 48.71 & 42.07 & 59.21 & 51.64 & 15.87 & 0.00 & 33.33 \\
\midrule
\multicolumn{10}{l}{\textit{Open-weight LLMs}} \\
Qwen3-1.7B & 36.31 & 44.11 & 25.22 & 22.07 & 39.58 & 23.77 & 6.35 & 0.00 & 11.11 \\
Qwen3-4B & 45.50 & 53.41 & 34.70 & 29.66 & 48.82 & 35.25 & 9.52 & 0.00 & 27.78 \\
Qwen3-8B & 47.46 & 54.92 & 38.58 & 28.28 & 51.33 & 33.90 & 9.68 & 0.00 & 14.52 \\
Qwen3-32B & 50.52 & 56.54 & 43.53 & 34.48 & 54.13 & 41.80 & 9.52 & 0.00 & 22.22 \\
Llama-3-8B-Inst & 24.12 & 29.30 & 17.89 & 11.03 & 26.54 & 14.75 & 1.59 & 0.00 & 5.56 \\
Llama-3-70B-Inst & 42.57 & 49.30 & 35.13 & 23.45 & 46.17 & 30.33 & 3.17 & 0.00 & 27.78 \\
DeepSeek-Coder-16B & 34.88 & 42.16 & 24.35 & 22.07 & 38.29 & 18.85 & 6.35 & 0.00 & 16.67 \\
DeepSeek-v4-flash & 50.20 & 56.65 & 41.59 & 36.55 & 53.30 & 43.44 & 15.87 & 8.33 & 16.67 \\
\bottomrule
\end{tabular}
\end{table*}

\begin{table*}[t]
\centering
\caption{Execution accuracy (\%) on T-SQL under the original BIRD difficulty and \textsc{UniQL} construction-source stratifications.}
\label{tab:appendix_difficulty_t_sql}
\scriptsize
\setlength{\tabcolsep}{2.2pt}
\begin{tabular}{lrrrrrrrrr}
\toprule
& & \multicolumn{3}{c}{BIRD difficulty} & \multicolumn{5}{c}{\textsc{UniQL} construction source} \\
\cmidrule(lr){3-5}\cmidrule(lr){6-10}
Model & All & Simple & Moderate & Challenging & Tool & LLM & Reflection & Rule Refinement & Human \\
\midrule
\multicolumn{10}{l}{\textit{Closed-weight LLMs}} \\
GPT-3.5-Turbo & 37.74 & 48.65 & 21.12 & 21.38 & 40.86 & 16.98 & 2.86 & 0.00 & 7.14 \\
GPT-5-mini & 53.85 & 60.00 & 44.18 & 45.52 & 56.21 & 38.68 & 22.86 & 0.00 & 57.14 \\
GPT-5.1-codex & 54.63 & 61.19 & 46.77 & 37.93 & 57.38 & 34.91 & 20.00 & 9.09 & 57.14 \\
Gemini-2.5-Pro & 55.61 & 62.16 & 46.77 & 42.07 & 58.92 & 37.74 & 11.43 & 0.00 & 21.43 \\
Claude-4.5-Sonnet & 56.26 & 62.27 & 48.49 & 42.76 & 59.36 & 37.74 & 17.14 & 9.09 & 28.57 \\
\midrule
\multicolumn{10}{l}{\textit{Open-weight LLMs}} \\
Qwen3-1.7B & 32.66 & 42.49 & 19.83 & 11.03 & 35.53 & 13.21 & 0.00 & 0.00 & 7.14 \\
Qwen3-4B & 46.41 & 55.78 & 34.27 & 25.52 & 49.56 & 26.42 & 11.43 & 0.00 & 14.29 \\
Qwen3-8B & 48.57 & 57.30 & 37.07 & 29.66 & 51.97 & 27.36 & 11.43 & 0.00 & 7.14 \\
Qwen3-32B & 51.69 & 58.92 & 41.81 & 37.24 & 54.61 & 37.74 & 11.43 & 0.00 & 14.29 \\
Llama-3-8B-Inst & 21.12 & 27.14 & 13.15 & 8.28 & 23.17 & 6.60 & 0.00 & 0.00 & 0.00 \\
Llama-3-70B-Inst & 40.61 & 48.65 & 30.60 & 21.38 & 44.15 & 16.04 & 0.00 & 0.00 & 14.29 \\
DeepSeek-Coder-16B & 31.68 & 40.32 & 20.69 & 11.72 & 34.58 & 11.32 & 0.00 & 0.00 & 7.14 \\
DeepSeek-v4-flash & 51.43 & 58.70 & 41.59 & 36.55 & 55.26 & 27.36 & 8.57 & 0.00 & 7.14 \\
\bottomrule
\end{tabular}
\end{table*}

\begin{table*}[t]
\centering
\caption{Execution accuracy (\%) on Teradata under the original BIRD difficulty and \textsc{UniQL} construction-source stratifications.}
\label{tab:appendix_difficulty_teradata}
\scriptsize
\setlength{\tabcolsep}{2.2pt}
\begin{tabular}{lrrrrrrrrr}
\toprule
& & \multicolumn{3}{c}{BIRD difficulty} & \multicolumn{5}{c}{\textsc{UniQL} construction source} \\
\cmidrule(lr){3-5}\cmidrule(lr){6-10}
Model & All & Simple & Moderate & Challenging & Tool & LLM & Reflection & Rule Refinement & Human \\
\midrule
\multicolumn{10}{l}{\textit{Closed-weight LLMs}} \\
GPT-3.5-Turbo & 23.86 & 30.92 & 12.28 & 15.86 & 39.49 & 8.51 & 1.38 & 0.00 & 4.92 \\
GPT-5-mini & 35.85 & 41.95 & 26.29 & 27.59 & 55.31 & 19.33 & 3.67 & 0.00 & 14.75 \\
GPT-5.1-codex & 35.07 & 42.05 & 24.78 & 23.45 & 54.47 & 18.04 & 3.21 & 0.00 & 16.39 \\
Gemini-2.5-Pro & 38.40 & 44.65 & 30.17 & 24.83 & 59.18 & 20.36 & 5.50 & 0.00 & 13.11 \\
Claude-4.5-Sonnet & 37.74 & 44.32 & 28.23 & 26.21 & 57.85 & 20.62 & 4.13 & 0.00 & 18.03 \\
\midrule
\multicolumn{10}{l}{\textit{Open-weight LLMs}} \\
Qwen3-1.7B & 22.43 & 30.16 & 11.42 & 8.28 & 37.92 & 6.19 & 0.92 & 0.00 & 6.56 \\
Qwen3-4B & 29.60 & 37.51 & 18.53 & 14.48 & 46.01 & 14.69 & 3.67 & 2.56 & 11.48 \\
Qwen3-8B & 29.34 & 36.54 & 20.04 & 13.10 & 45.65 & 13.32 & 2.93 & 3.33 & 15.91 \\
Qwen3-32B & 32.27 & 39.68 & 21.55 & 19.31 & 49.03 & 17.78 & 5.50 & 2.56 & 11.48 \\
Llama-3-8B-Inst & 15.45 & 20.76 & 6.68 & 9.66 & 26.45 & 3.61 & 0.92 & 0.00 & 3.28 \\
Llama-3-70B-Inst & 26.47 & 32.86 & 17.67 & 13.79 & 43.48 & 10.05 & 0.92 & 0.00 & 8.20 \\
DeepSeek-Coder-16B & 19.95 & 26.92 & 9.70 & 8.28 & 33.82 & 5.67 & 0.92 & 0.00 & 3.28 \\
DeepSeek-v4-flash & 30.90 & 37.62 & 20.91 & 20.00 & 48.91 & 13.92 & 3.67 & 0.00 & 11.48 \\
\bottomrule
\end{tabular}
\end{table*}

\begin{table*}[t]
\centering
\caption{Execution accuracy (\%) on Trino under the original BIRD difficulty and \textsc{UniQL} construction-source stratifications.}
\label{tab:appendix_difficulty_trino}
\scriptsize
\setlength{\tabcolsep}{2.2pt}
\begin{tabular}{lrrrrrrrrr}
\toprule
& & \multicolumn{3}{c}{BIRD difficulty} & \multicolumn{5}{c}{\textsc{UniQL} construction source} \\
\cmidrule(lr){3-5}\cmidrule(lr){6-10}
Model & All & Simple & Moderate & Challenging & Tool & LLM & Reflection & Rule Refinement & Human \\
\midrule
\multicolumn{10}{l}{\textit{Closed-weight LLMs}} \\
GPT-3.5-Turbo & 35.85 & 45.51 & 21.34 & 20.69 & 38.56 & 18.07 & 7.84 & 0.00 & 0.00 \\
GPT-5-mini & 49.02 & 56.11 & 39.66 & 33.79 & 50.91 & 44.58 & 23.53 & 0.00 & 11.76 \\
GPT-5.1-codex & 50.72 & 58.16 & 41.81 & 31.72 & 53.16 & 37.35 & 23.53 & 0.00 & 17.65 \\
Gemini-2.5-Pro & 55.61 & 61.62 & 47.63 & 42.76 & 58.10 & 45.78 & 25.49 & 0.00 & 11.76 \\
Claude-4.5-Sonnet & 56.06 & 62.27 & 48.06 & 42.07 & 58.10 & 46.99 & 33.33 & 0.00 & 23.53 \\
\midrule
\multicolumn{10}{l}{\textit{Open-weight LLMs}} \\
Qwen3-1.7B & 34.16 & 43.46 & 22.20 & 13.10 & 36.89 & 13.25 & 7.84 & 0.00 & 5.88 \\
Qwen3-4B & 41.46 & 51.57 & 28.66 & 17.93 & 44.59 & 19.28 & 9.80 & 0.00 & 5.88 \\
Qwen3-8B & 44.20 & 53.08 & 33.62 & 21.38 & 46.99 & 25.30 & 17.65 & 0.00 & 5.88 \\
Qwen3-32B & 46.61 & 54.27 & 38.58 & 23.45 & 48.87 & 36.14 & 17.65 & 0.00 & 17.65 \\
Llama-3-8B-Inst & 22.75 & 28.43 & 15.95 & 8.28 & 24.69 & 7.23 & 5.88 & 0.00 & 0.00 \\
Llama-3-70B-Inst & 39.05 & 48.22 & 28.66 & 13.79 & 42.12 & 15.66 & 5.88 & 16.67 & 11.76 \\
DeepSeek-Coder-16B & 32.53 & 41.08 & 21.77 & 12.41 & 35.29 & 8.43 & 9.80 & 0.00 & 5.88 \\
DeepSeek-v4-flash & 47.59 & 55.68 & 37.93 & 26.90 & 49.75 & 32.53 & 25.49 & 16.67 & 23.53 \\
\bottomrule
\end{tabular}
\end{table*}

\clearpage
\section{Artifact License and Intended Use}
\label{app:artifact_license}

\textsc{UniQL} is constructed from the public BIRD development set~\cite{li2023can}. 
We use BIRD as a research benchmark for text-to-SQL evaluation and extend it by constructing executable SQL annotations across multiple SQL dialects. 
This use is consistent with the original purpose of BIRD as a benchmark for evaluating and advancing text-to-SQL systems in research settings.

We plan to release the \textsc{UniQL} benchmark, construction metadata, evaluation scripts, and accompanying documentation to support reproducibility and future research. 
The benchmark is intended for evaluating cross-dialect text-to-SQL systems, analyzing SQL dialect robustness, and reproducing or extending the experiments reported in this paper. 
Since \textsc{UniQL} is a derivative benchmark based on BIRD, users of \textsc{UniQL} are required to comply with the license, citation requirements, and usage terms of the source BIRD dataset. 
In particular, the released benchmark data should be used only in contexts that are compatible with the original access conditions and intended research use of BIRD.

For artifacts created by us, including evaluation scripts, construction utilities, and documentation, we will specify the corresponding license in the released repository. 
The release will also document the intended use of each artifact, citation requirements, and any restrictions inherited from the source benchmark.

\section{Data Privacy and Content Safety}
\label{app:data_safety}

\textsc{UniQL} does not involve collecting new private user data. 
The natural language questions, database schemas, database contents, evidence fields, and original SQLite SQL queries are derived from the public BIRD development set~\cite{li2023can}. 
Our construction process focuses on translating and verifying SQL annotations across target dialects, and does not introduce new personal identifiers or user-generated private content.

We reviewed the data used in \textsc{UniQL}, including natural language questions, SQL annotations, evidence fields, and database contents, for personally identifying information and offensive content. 
The dialect translation process only rewrites SQL queries into target SQL dialects and preserves the public benchmark setting of the source dataset. 
Any released artifacts will follow the usage terms of BIRD and will be distributed with documentation describing their intended research use and inherited restrictions.

\end{document}